\documentclass[lettersize,journal]{IEEEtran}
\usepackage{amsmath,amsfonts}
\usepackage{algorithmic}
\usepackage{algorithm}
\usepackage{array}
\usepackage[caption=false,font=normalsize,labelfont=sf,textfont=sf]{subfig}
\usepackage{textcomp}
\usepackage{stfloats}
\usepackage{url}
\usepackage{verbatim}
\usepackage{graphicx}
\usepackage{cite}
\usepackage{multirow}
\usepackage{xcolor}
\usepackage{amsmath} 

\DeclareMathOperator*{\argmax}{argmax}

\usepackage{caption}

\makeatletter
\let\NAT@parse\undefined
\makeatother
\usepackage{hyperref}  

\begin{document}

\title{CoreEditor: Correspondence-constrained Diffusion for Consistent 3D Editing}

\author{Zhe Zhu, Honghua Chen, Peng Li, and Mingqiang Wei,~\IEEEmembership{Senior Member,~IEEE}

\thanks{
    \IEEEcompsocthanksitem Z. Zhu, P. Li, and M. Wei are with the School of Computer Science and Technology, Nanjing University of Aeronautics and Astronautics, Nanjing, China; and also with the Shenzhen Research Institute, Nanjing University of Aeronautics and Astronautics, Shenzhen, China  (e-mail: zhuzhe0619@nuaa.edu.cn; pengl@nuaa.edu.cn; mingqiang.wei@gmail.com).
    \IEEEcompsocthanksitem H. Chen is with the School of Data Science, Lingnan University, Hong Kong, China (e-mail: honghuachen@LN.edu.hk).
    \IEEEcompsocthanksitem Corresponding author: H. Chen.
}
}

\maketitle

\markboth{Journal of \LaTeX\ Class Files,~Vol.~14, No.~8, August~2021}%
{Shell \MakeLowercase{\textit{et al.}}: A Sample Article Using IEEEtran.cls for IEEE Journals}

\begin{abstract}
Text-driven 3D editing is an emerging task that focuses on modifying scenes based on text prompts. Current methods often adapt pre-trained 2D image editors to multi-view observations, using specific strategies to combine information across views. 
However, these approaches still struggle with ensuring consistency across views, as they lack precise control over the sharing of information, resulting in edits with insufficient visual changes and blurry details.
In this paper, we propose CoreEditor, a novel framework for consistent text-to-3D editing. 
At the core of our approach is a novel correspondence-constrained attention mechanism, which 
enforces structured interactions between corresponding pixels that are expected to remain visually consistent during the diffusion denoising process.
Unlike conventional wisdom that relies solely on scene geometry, we enhance the correspondence by incorporating semantic similarity derived from the diffusion denoising process. This combined support from both geometry and semantics ensures a robust multi-view editing process.
Additionally, we introduce a selective editing pipeline that enables users to choose their preferred edits from multiple candidates, creating a more flexible and user-centered 3D editing process. 
Extensive experiments demonstrate the effectiveness of CoreEditor, showing its ability to generate high-quality 3D edits, significantly outperforming existing methods.
Our code is available at \emph{\textcolor{magenta}{https://github.com/czvvd/CoreEditor}}.
\end{abstract}


\IEEEdisplaynontitleabstractindextext

%
\IEEEpeerreviewmaketitle

\section{Introduction}
\label{sec:intro}
\begin{figure*}[h]
  \centering
  \includegraphics[width=\textwidth]{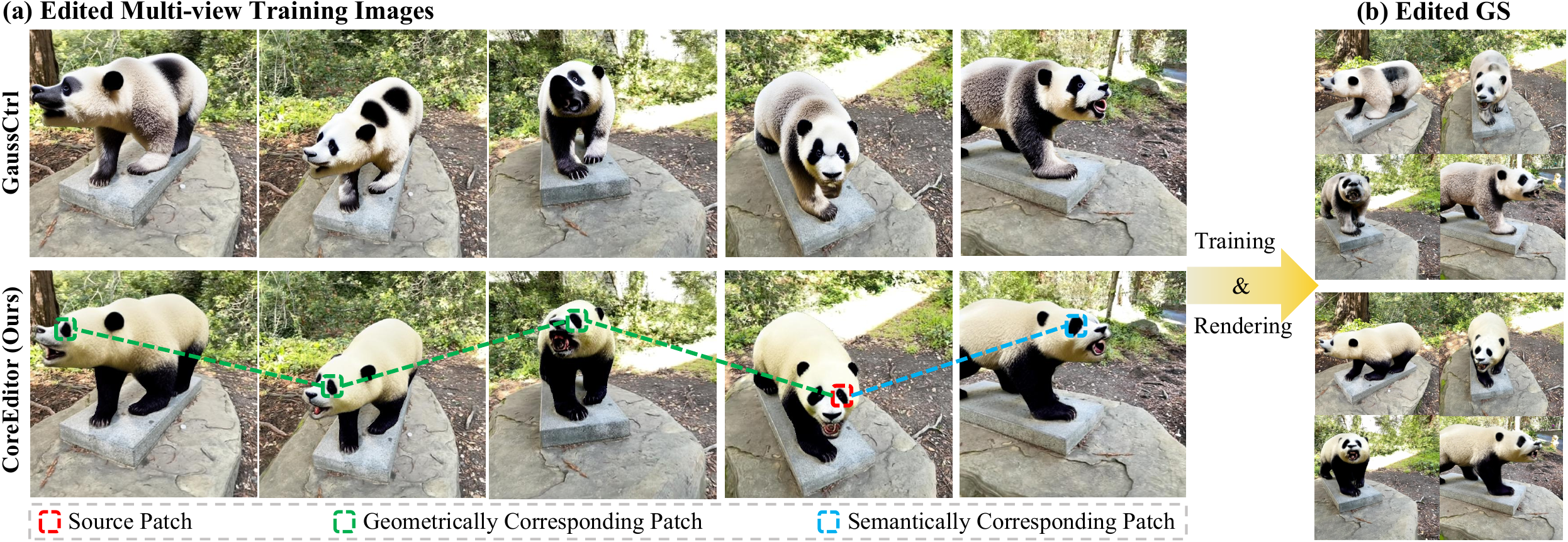}
\caption{\textbf{Key features of our method and visual comparison with the recent GaussCtrl~\cite{gaussctrl} method.}
(a) Visual comparison of edited multi-view training images. CoreEditor integrates geometric and semantic correspondences into the T2I diffusion model, ensuring 3D-consistent edits.
(b) Visual comparison of rendered edited results. With consistent multi-view images, CoreEditor generates results with sharper textures.}

  \label{fig:teaser}
\end{figure*}

Recent years have witnessed remarkable progress in neural 3D representations, with pioneering works like NeRF~\cite{nerf} and Gaussian Splatting~\cite{3dgs} achieving photorealistic novel view synthesis.
These methods excel at reconstructing high-quality 3D scenes from multi-view images. However, once constructed, modifying these scenes to align with user preferences remains a challenging task. Developing such a 3D editing tool has become a critical research focus. 

With recent successes of diffusion-based text-to-image (T2I) models~\cite{ddpm,latentdiffusion,controlnet}, most text-driven 3D editing methods typically leverage these T2I models to edit multi-view images in a zero-shot manner. Yet, the stochastic nature of diffusion models often leads to inconsistent editing results across views, producing imprecise 3D editing with blurry textures.
To overcome the inconsistency issue, the pioneering InstructNerf2Nerf~\cite{instructnerf2nerf} proposes an iterative dataset update scheme but suffers from slow optimization.
In an effort to accelerate the editing process, some recent works~\cite{rojas2025datenerf,gaussctrl,dge} jointly edit multiple views and align the editing pattern between these views through various strategies, such as depth-conditioned ControlNet~\cite{controlnet}, cross-frame attention~\cite{tuneavideo}, and cross-view feature interpolation~\cite{dge}. 
However, these strategies lack precise constraints on the direction of multi-view information exchange, thereby compromising the consistency of local image details, especially under significant viewpoint variations and complex 3D scene occlusions.
As presented in Fig.~\ref{fig:teaser} (a), GaussCtrl~\cite{gaussctrl} produces inconsistent and low-quality multi-view edits, resulting in results with blurry texture.

In this paper, we propose CoreEditor to address these issues. 
CoreEditor achieves 3D consistency by integrating precise multi-view constraints into a pre-trained T2I diffusion model.
It has been shown that, in a T2I model, tokens of different images can collaborate with each other in the attention module while still generating reasonable results~\cite{pnp,masactrl,flatten,dge,tuneavideo,dreammatcher,Zhou2024storydiffusion}. 
Enlightened by this, our key idea is introducing a Correspondence-constrained Attention (CCA) in the diffusion U-Net, where image patches rendered from the same 3D point are constrained to interact with one another to improve visual consistency. 
Without the need for fine-tuning or re-training the diffusion model, the revised information flow direction of the attention module can significantly improve the consistency between multi-view generated content.

Despite its effectiveness, we identify two key challenges when directly applying CCA to diverse 3D editing:
(1) When the camera distances between views are large, especially in some $360^\circ$ scenes, the background image patches may have few geometrically corresponding patches in other views due to occlusion. 
The insufficient token count causes the attention process to become highly unstable, often leading to low-quality and over-saturated outputs (see Fig.~\ref{fig:ablationSemantic}).
(2) When there is a significant disparity among per-view editing results, CCA tends to ``average'' these edits, resulting in unnatural edits.

To address the first one, we design a geometric and semantic co-supported approach to extract the multi-view correspondences. The key insight here is that semantically similar patches can also be involved in attention to improve consistency. As shown in Fig.~\ref{fig:teaser} (a), although the left eye is occluded in the rightmost image, the accessible right eye is also expected to be visually consistent with the left one. Therefore, we are inspired to enrich the sparse geometric correspondences with semantic information. Specifically, in regions where geometric correspondences are unavailable, additional correspondences are calculated based on the diffusion feature similarity. With this comprehensive correspondence, CCA generates more plausible multi-view edits.
Regarding the second problem, we introduce a selective editing pipeline, where users are allowed to select their preferred editing pattern from the per-view editing results. The selected edit is then injected into the diffusion model by a Reference Attention (RA).
This approach ensures preliminary alignment of the global editing style, enabling CCA to focus solely on local consistency. Moreover, CoreEditor can generate diverse yet faithful 3D edits by selecting different per-view editing patterns.

We conduct comprehensive experiments including various scenes and editing prompts. The results demonstrate that CoreEditor achieves superior results than its competitors in terms of multi-view consistency and editing quality.
The main contributions of this work can be summarized as follows.
\begin{itemize}
\item We design a novel 3D editing method called CoreEditor, which significantly improves the multi-view consistency by a Correspondence-constrained Attention mechanism.
\item We propose a geometric and semantic co-supported approach to build the multi-view correspondences, which significantly improves editing quality in complex scenes.
\item We introduce a selective editing pipeline, which allows a flexible and user-centered 3D editing experience.
\end{itemize}

\section{Related Work}
\label{sec:related}
\subsection{Text-driven 3D Editing}
Early methods~\cite{text2mesh, clipnerf, nerfart} primarily leverage vision-language models~\cite{clip} for text-driven 3D stylization. However, their capabilities are often limited to modifying only the global style of the scene.
Building upon the success of diffusion models, DreamFusion~\cite{dreamfusion} introduces a score distillation sampling (SDS) loss for 3D generation from arbitrary text, which implicitly transfers prior knowledge from a pre-trained T2I model. The SDS loss has since been applied to 3D generation and editing in several subsequent works~\cite{voxe, dreameditor, tipeditor, watchyourstep, chen2024mvip,dds,pds}.

InstructNerf2Nerf~\cite{instructnerf2nerf} is the first method to explicitly utilize a T2I model for this task. It addresses the issue of multi-view inconsistency by iteratively alternating between editing the training images and optimizing the 3D scene, which, however, results in a slow editing process.
Following-up works improve the editing performance and speed by leveraging the explicit properties of 3D Gaussian splitting~\cite{gaussianeditor1, gaussianeditor2, vcedit,in2024editsplat}, latent space optimization~\cite{shapeditor, freditor, latenteditor}, personalized editing~\cite{tipeditor,gaussedit,he2024customize}, progressive editing~\cite{proedit}, and 3D-aware fine-tuning of the diffusion model~\cite{luo20253denhancer,cai2024mv2mv,chen2024consistdreamer,wynn2025morpheus}.
Depth images are frequently used in recent methods to link different views. For example, VICA-NeRF~\cite{vica} and DATENeRF~\cite{rojas2025datenerf} project edited images to other views using depth information.
Among them, a prevalent approach involves propagating information between different views during multi-view joint editing. 
GaussCtrl~\cite{gaussctrl} utilizes depth as guidance for ControlNet~\cite{controlnet} and aligns the latent code of different views. DGE~\cite{dge} applies cross-frame attention to edit key views, blending the edited features based on epipolar constraints. EditSplat~\cite{in2024editsplat} proposes a multi-view classifier-free guidance strategy to guide the diffusion model.
InterGSEdit~\cite{wen2025intergsedit} enforces multi-view consistency through a dynamic gating mechanism in cross-view attention; however, it still relies on a set of edited key frames, which can be inconsistent.
Despite these advances, existing strategies often lack accurate constraints during the multi-view joint editing. Thereby, they typically fail to maintain precise 3D consistency. In contrast, we propose a novel correspondence-constrained attention mechanism, where only the corresponding tokens across views can communicate with each other. This enables a reliable information exchange across views, minimizing the introduction of irrelevant content and ensuring precise 3D consistency.

\begin{figure*}[h]
  \centering
  \includegraphics[width=\textwidth]{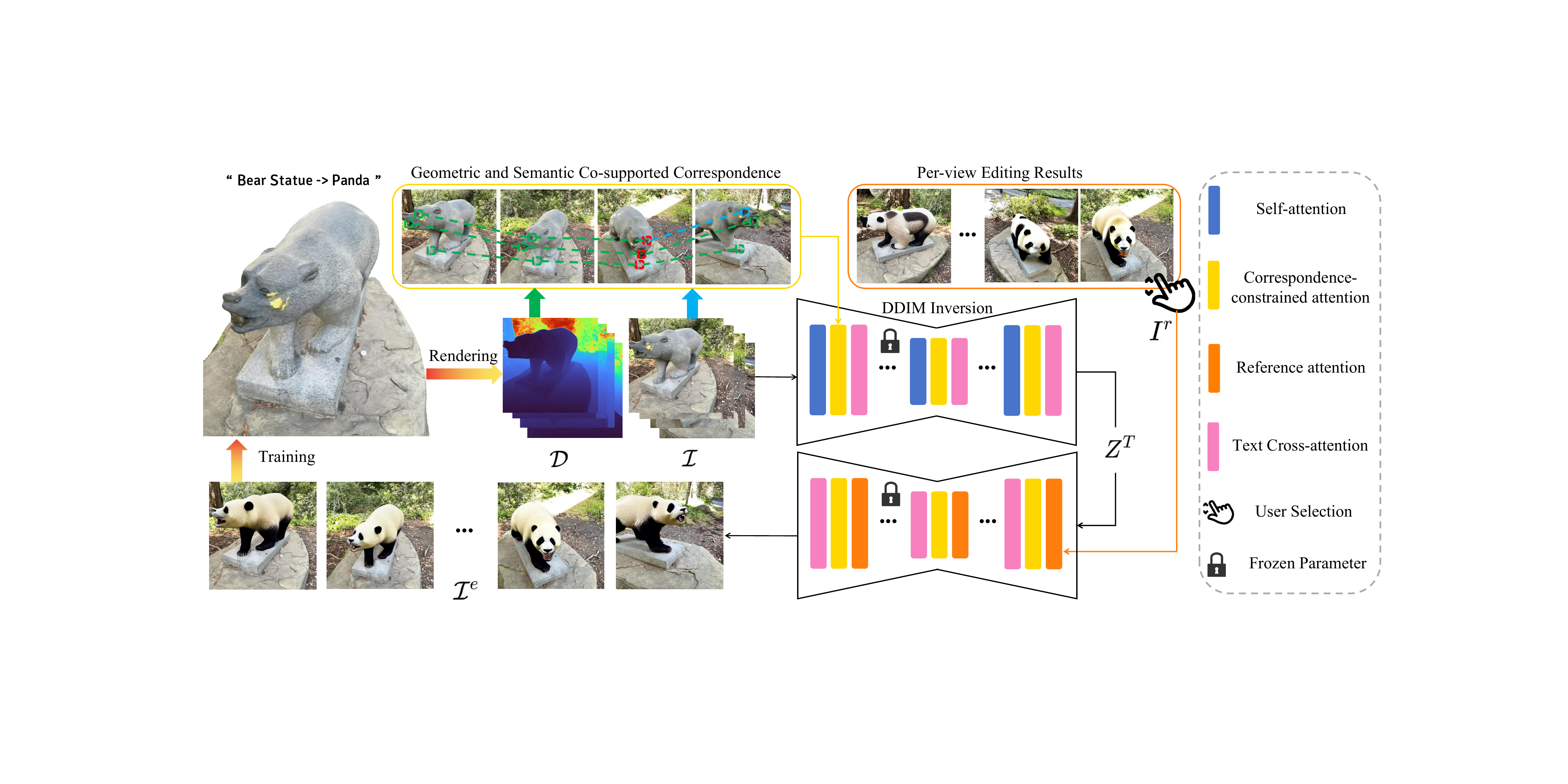}
\caption{\textbf{Overview of CoreEditor.} Our method edits the rendered multi-view images ($\mathcal{I}$) into a consistent image set $\mathcal{I}^e$, which is then used to update the original GS model. The process ensures 3D consistency through two key steps:
(1) Once the user selects a preferred edit, $I^r$, we integrate its pattern into the diffusion model using Reference Attention.
(2) After the geometry and semantic co-supported correspondence set has been established, we inject it into the diffusion model by Correspondence-constrained Attention.}

  \label{fig:overview}
\end{figure*}

\subsection{3D-aware Diffusion Model}
Extensive efforts have been made to introduce 3D awareness into pre-trained T2I diffusion models, transforming them into multi-view generators. The pioneering work Zero-1-to-3~\cite{zero123} incorporates relative pose as an additional condition to generate novel views from a single-view observation. Building on this, SyncDreamer~\cite{syncdreamer} connects corresponding pixels through a feature volume. MVDream~\cite{shi2024mvdream} and Imagedream~\cite{imagedream} process all pixels across multi-view images together, significantly increasing computational complexity. To allow for precise control and reduce computational costs, recent methods integrate 3D constraints~\cite{epidiff,spad,mvdiffusion,li2024era3d,huang2024mv} and camera positional embedding~\cite{eschernet} into the attention module.
Our CoreEditor shares similarities with these approaches in extending a pre-trained diffusion model to a multi-view network. However, unlike these methods, which introduce a large number of trainable parameters and require extensive fine-tuning, our method can be seamlessly integrated into existing diffusion models in a zero-shot manner.

\section{Method}
\label{sec:method}

\subsection{Preliminaries}
\noindent\textbf{3D Gaussian Splatting.}
Our method adopts Gaussian Splatting (GS)~\cite{3dgs} as the 3D representation. In GS, a 3D scene is represented as a collection of Gaussian primitives, each characterized by its center coordinate \( \mu \), covariance matrix \( \Sigma \), opacity \( \sigma \), and color \( c \) represented by spherical harmonic coefficients. To enable real-time rendering, GS employs a splatting rendering approach, where the color is computed by blending the contributions of Gaussians projected onto that pixel. Similar to NeRF~\cite{nerf}, GS is also capable of reconstructing depth by computing the weighted average of the distance values of the projected Gaussians.

\noindent\textbf{Latent Diffusion Model.}
Recently, latent diffusion model~\cite{latentdiff} has emerged as a dominant architecture for image generation. It reduces computational overhead by compressing images into a low-dimensional latent space, where both the forward and backward diffusion processes\cite{ddpm,ddim} are performed. The denoising network employs a U-Net~\cite{unet} architecture, with each layer consisting of a self-attention (SA) and a text cross-attention (CA) module.

\noindent\textbf{DDIM Inversion for Image Editing.}
DDIM inversion~\cite{ddim} enables the reversal of an image to its corresponding noise representation in diffusion space. A typical image editing workflow involves first inverting the image to noise $Z^T$, then regenerating the edited version using the inverted noise and a target text prompt. To preserve the original layout, additional constraints, such as attention feature replacement~\cite{pnp,p2p}, are often applied during the editing process. 

\subsection{Overview: Selective Editing Pipeline}
Given a 3D GS model $\mathcal{G}$ and a text prompt $T$, we propose CoreEditor to modify $\mathcal{G}$ such that it faithfully aligns with $T$. Ideally, if multi-view consistent edited images can be obtained, $\mathcal{G}$ can be updated accordingly to achieve high-quality 3D edits.
As shown in Fig.~\ref{fig:overview}, CoreEditor ensures consistent multi-view editing through a re-designed denoising U-Net architecture. Specifically, after rendering multi-view source images $\mathcal{I}=\{I_i\}^N_{i=1}$ and depth maps $\mathcal{D}=\{D_i\}^N_{i=1}$ from $N$ views, CoreEditor performs the editing in two main steps:

(1) 
We first align multi-view edits towards a user-selected style.
In particular, each image in $\mathcal{I}$ is firstly edited using a standard inversion-based approach, during which we save the intermediate diffusion features at each layer. After the editing process, users can select their preferred result, $I^r$, which serves as the reference edit for the following steps. The corresponding feature $F^r$ is then injected into the subsequent steps through Reference Attention (RA) (Sec.~\ref{sec:refattn}).

(2) In the second step, we incorporate multi-view constraints into the diffusion process and jointly edit images in $\mathcal{I}$ to an image set $\mathcal{I}^e$ with consistent local details. In detail, $\mathcal{I}$ and $\mathcal{D}$ are used to build geometric and semantic co-supported correspondence (Sec.~\ref{sec:corr}). Those correspondences are integrated by introducing a Correspondence-constrained Attention (CCA) module (Sec.~\ref{sec:corrattn}) in the U-Net. With the modified diffusion model, we use the inversion-based editing method to get $\mathcal{I}^e$, which is then used to optimize $\mathcal{G}$.

\textbf{During the above process, the diffusion model is kept frozen, without introducing any additional training}.

\begin{figure}
    \centering
    \includegraphics[width=1\linewidth]{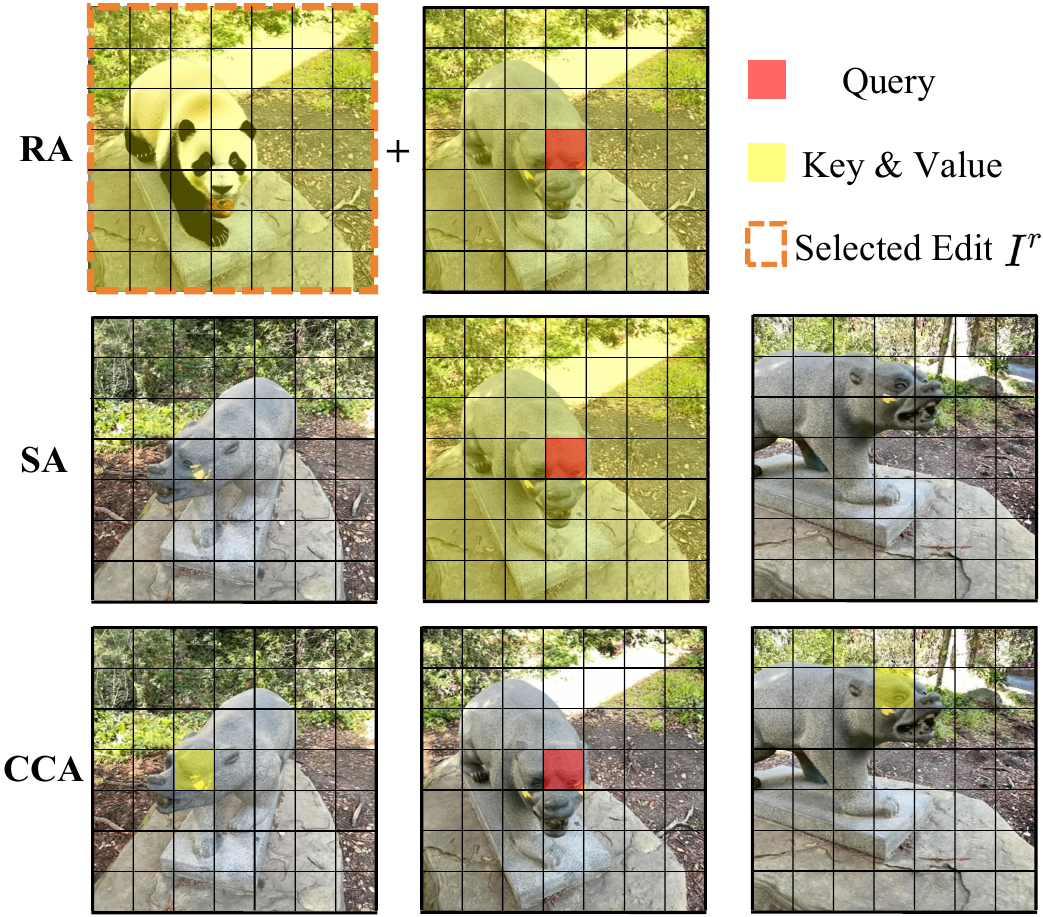}
    \caption{Difference between the calculation of RA, SA, and CCA. Compared with the original SA, RA regards the selected edit as an additional set of key and value. To improve local consistency, CCA enforces an image patch token to only interact with the corresponding patches in other views.}
    \label{fig:attn}
\end{figure}

\subsection{Reference Attention (RA)}
\label{sec:refattn}
Since edited results for the same prompt can be completely different across views, posing difficulties in producing high-quality edits while keeping consistency between them, we allow the users to select their preferred editing pattern, which guides the entire editing process through attention feature injection. 
Specifically, we modify the SA module within the diffusion backward process, transforming it into an RA module. As presented in Fig.~\ref{fig:attn}, compared with SA, the diffusion feature $F^r$ of $I^r$ serves as an additional key and value in RA, thus facilitating the alignment of the editing style. Given the multi-view input features $\mathcal{F} = \{F_i\}_{i=1}^{N}$ in the reference attention module, the output feature $Z{i}$ for the $i$-th view is computed as follows:

\begin{equation}
\label{eqn:refattn}
\begin{split}
Z_{i} &= \lambda \cdot \text{softmax}\left(\frac{W_qF_{i}(W_kF^r)^\top}{\sqrt{d_k}}\right)W_vF^r
\\& + (1-\lambda) \cdot \text{softmax}\left(\frac{W_qF_{i}(W_kF_{i})^\top}{\sqrt{d}}\right)W_vF_{i}
\end{split}
\end{equation}

where $W_q$, $W_k$, and $W_v$ are projection matrices in the attention module,  and $\lambda \in [0, 1]$ is a coefficient that modulates the weighting of the reference and original attention terms.
After injecting $I^r$, the global editing patterns have been aligned, significantly reducing the solution space for consistent results.
\textbf{Furthermore, manual selection can also be automated through the human preference predictor~\cite{xu2023imagereward}, establishing a fully automatic workflow (See results in Sec.~\ref{sec:abla}).}

\subsection{Geometric and Semantic Co-supported Correspondence}
\label{sec:corr}
We first build image correspondence relationships between views to serve as precise 3D constraints for the diffusion model. 
For a pixel coordinate $P=(x_s, y_s)$ in the $s$-th view, our goal is to find its correspondences $\mathcal{C} = \left\{ (x_i, y_i) \mid i = 1, 2, \dots, N, \, i \neq s \right\}$ in the remaining $N-1$ views. Geometric correspondence can be directly derived from the depth maps $\mathcal{D}$. ‌The geometrically corresponding pixel $(x_a, y_a)$ of $P$ in a target view $a$ is obtained as:

\begin{equation}
    (x_a, y_a)=\text{Proj}(\text{BackProj}((x_s, y_s), D_s, K, E_s), K, E_a)
\end{equation}
where $K$, $E_s$, and $E_a$ are the intrinsic and extrinsic camera parameters of views $s$ and $a$. In this process, $P$ is first back-projected into 3D space using the depth $D_s$ of this view, then re-projected to the target view $a$ to obtain its corresponding pixel location.
To handle potential occlusions, we compute a re-projection error to construct a correspondence mask $M = \{ m_{i} \mid m_{i} \in \{0, 1\}, \, i = 1, 2, \dots, N, \, i \neq s \}$, filtering out unreliable matches.

However, as discussed in Sec.~\ref{sec:intro}, certain pixels should exhibit visual consistency across views even without valid geometric correspondence. These implicit relationships can enrich the sparse geometric correspondence and stabilize the attention mechanism.
Inspired by recent work demonstrating emergent correspondence in image diffusion models~\cite{dift}, we obtain additional semantic correspondence based on diffusion features. To be specific, we apply a single forward and backward step of DDPM to $\mathcal{I}$ and extract the last-layer feature maps $\mathcal{H}=\{H_i\}^N_{i=1}$ output by the U-Net. 
Then, for a target view $b$ lacking valid geometric correspondence for $P$, We define its corresponding pixel $(x_b, y_b)$ as the location with the highest feature cosine similarity:
\begin{equation}
    (x_b, y_b)= \argmax\limits_{(x, y)} \frac{H_s(x_s, y_s)
    \cdot H_b(x, y)}{\|H_s(x_s, y_s)\|     
     \|H_b(x, y)\|}
\end{equation}
Since valuable information exists only in some specific target views, we only use semantic correspondences with the highest cosine similarity exceeding a threshold value $\beta$. Other semantic correspondences remain masked. Here, to avoid additional hyper-parameter tuning, we typically set $\beta$ to 0.9, which performs well across various scenes.
Finally, a comprehensive correspondence set $\mathcal{C}$ has been established, accompanied by a mask $M$ indicating the validity of each correspondence.

\begin{figure*}[h]
  \centering
  \includegraphics[width=\textwidth]{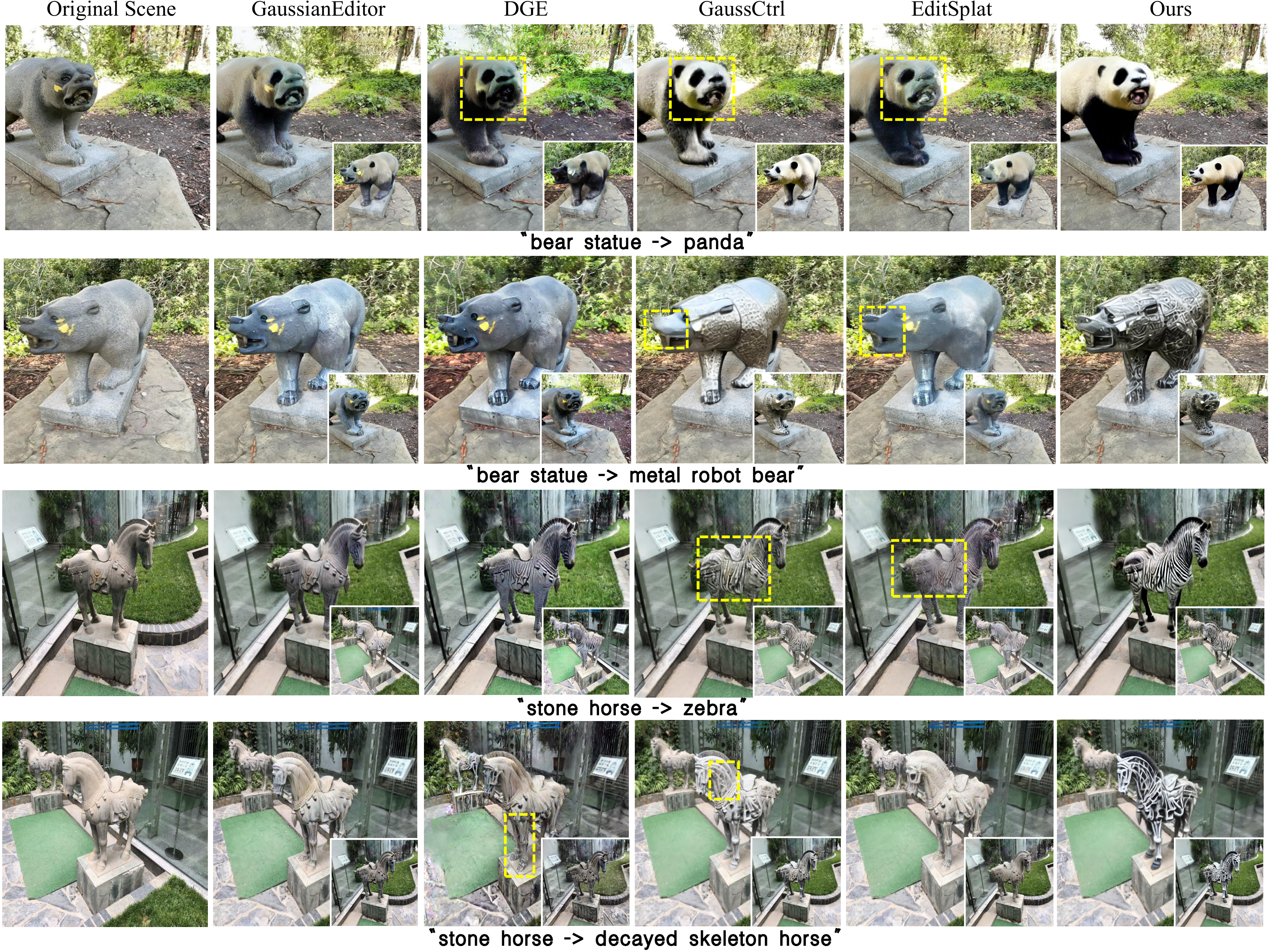}
\caption{Visual comparison with state-of-the-art methods~\cite{gaussianeditor2,dge,gaussctrl,in2024editsplat} in the ``bear'' and ``stone horse'' scenes. We provide results rendered from two views for each edited scene. Blurry regions are highlighted with yellow dash boxes.
}
  \label{fig:comp}
\end{figure*}

\subsection{Correspondence-constrained Attention (CCA)}
\label{sec:corrattn}
We then incorporate $\mathcal{C}$ into the diffusion U-Net through the proposed CCA module. 
Specifically, in both the DDIM inversion and the denoising backward process, a CCA is placed after each self-attention/reference attention module.  
Given the multi-view features $\mathcal{Z}=\{Z_i\}^N_{i=1}$ output by self/reference attention modules, the output token $P^{'}$ from CCA can be calculated as follows:
\begin{equation}
\begin{split}
Q &= Z_s(x_s,y_s), \\
K=V &= \{Z_i(\mathcal{C}[i])\mid i=1,2,\dots,N\}, \\
P' &= \text{softmax}\left(\frac{QK^\top}{\sqrt{d}} + M'\right)V
\end{split}
\end{equation}

Here, we do not introduce additional parameters for re-projecting the latent; rather, we alter the direction of the information flow and perform the attention calculation once more.
Fig.~\ref{fig:attn} illustrates how CCA works: It constrains $P$ to only interact with image tokens belonging to $\mathcal{C}$ instead of querying image tokens within the source view. 
Meanwhile, we filter out correspondences that are identified as invalid in Sec.~\ref{sec:corr} by extending $M$ to an attention mask $M^{'}$.
Without any need for fine-tuning the diffusion model, the precise interaction between views significantly improves the multi-view consistency, thereby making the diffusion model a high-quality 3D editor.

\section{Experiment}

\begin{figure*}[h]
  \centering
  \includegraphics[width=\textwidth]{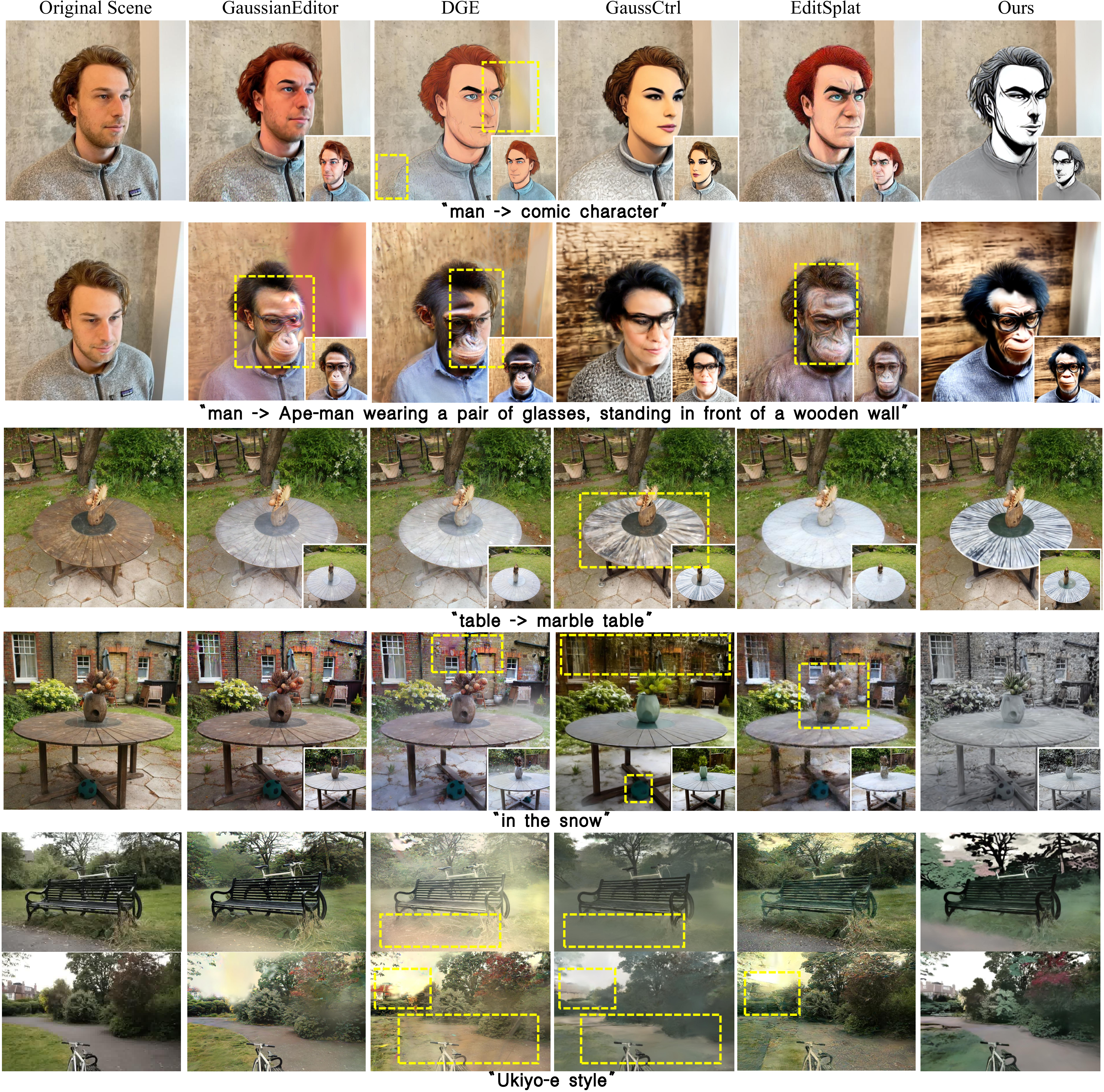}
\caption{Visual comparison with state-of-the-art methods~\cite{gaussianeditor2,dge,gaussctrl,in2024editsplat} in the ``face'', ``garden'', and ``bicycle'' scenes.  We provide results rendered from two views for each edited scene. Blurry regions are highlighted with yellow dash boxes.
}
  \label{fig:comp2}
\end{figure*}

\subsection{Experimental Setup}

\noindent\textbf{Implementation Details.}
We adopt Splatfacto as the 3D representation, a modified version of Gaussian Splatting~\cite{3dgs}, implemented within the Nerfstudio library~\cite{nerfstudio}. For the text-to-image (T2I) model, we use Stable Diffusion v1.5~\cite{latentdiff} combined with its corresponding depth-conditioned ControlNet~\cite{controlnet}, implemented in the Diffusers library~\cite{diffusers}.
For local editing, where the irrelevant background is expected to remain unchanged, we utilize Lang-SAM~\cite{SAM} to generate masks for the edited images, filtering out the background. 
In all the experiments, we manually select $I^r$ from the per-view edits.
Then, the proposed multi-view editing is applied with 500 steps of GS optimization. 
The commonly used L1 and LPIPS~\cite{lpips} losses are applied as the objective function during the GS optimization.
The number of multi-view images $N$ is set to $20$, randomly sampled in the training set. During the DDIM inversion and backward denoising process, we use $20$ diffusion steps.
For scene stylization prompts, $\lambda$ is set to 0.3. For other prompts, such as human character editing, which exhibit high variability and may generate vastly different styles, we employ a higher $\lambda$ value of 0.5. This ensures that the selected edit is consistently replicated across different views.
For all scenes, $\beta$ is set as $0.9$.

\noindent\textbf{Data.}
We evaluate CoreEditor on seven scenes from InstructNeRF2NeRF~\cite{instructnerf2nerf}, Mip-NeRF 360~\cite{mipnerf360}, BlendMVS~\cite{blendedmvs}, and NerfStudio~\cite{nerfstudio},
including the ``bear'' and ``face'' scenes from \cite{instructnerf2nerf}, the ``bicycle'' and ``garden'' scenes from \cite{mipnerf360}, the ``stone horse'' and ``dinosaur'' scenes from \cite{blendedmvs}, and the ``dozer'' scene from \cite{nerfstudio}. 
The performance of our method is evaluated across a total of $20$ challenging prompts, covering tasks such as local editing, global stylization, and human character modification.

\begin{table*}[h]
    \renewcommand\arraystretch{1.2}
    \centering
    \caption{Quantitative comparison with recent methods~\cite{gaussianeditor2,dge,gaussctrl,in2024editsplat} (CLIP$_{sim}:$ CLIP similarity scores, CLIP$_{dir}:$ CLIP directional similarity scores, Met3R values~\cite{asim2025met3r}, and User study voting rates). For CLIP-based metrics, we report both the mean and median values across all evaluation cases (shown as mean / median).}
    \label{tab:comp}
    \small
    \begin{tabular}{|c|c|c|c|c|c|}
        \hline
        \multirow{2}{*}{Methods} & \multicolumn{2}{c|}{CLIP Metrics} & \multirow{2}{*}{Met3R$\downarrow$} & \multicolumn{2}{c|}{User Study} \\
        \cline{2-3} \cline{5-6}
        & CLIP$_{sim}\uparrow$ & CLIP$_{dir}\uparrow$ & & Quality$\uparrow$ & Consistency$\uparrow$ \\
        \hline
        GaussianEditor~\cite{gaussianeditor2} & 0.244 / 0.235 & 0.086 / 0.081 & - & 7.0\% & 8.8\% \\
        DGE~\cite{dge} & 0.259 / 0.242 & 0.123 / 0.137 & 0.390 & 14.6\% & 15.0\% \\
        GaussCtrl~\cite{gaussctrl} & 0.257 / 0.251 & 0.128 / 0.126 & 0.372 & 16.2\% & 14.8\% \\
        EditSplat~\cite{in2024editsplat} & 0.261 / 0.252 & 0.130 / 0.132 & 0.336 & 17.0\% & 19.4\% \\
        Ours & \textbf{0.270} / \textbf{0.259} & \textbf{0.145} / \textbf{0.141} & \textbf{0.281} & \textbf{45.2\%} & \textbf{42.0\%} \\
        \hline
    \end{tabular}
\end{table*}

\noindent\textbf{Baselines.}
We compare CoreEditor against four state-of-the-art GS-based 3D editing methods: GaussianEditor~\cite{gaussianeditor2}, DGE~\cite{dge}, GaussCtrl~\cite{gaussctrl}, and EditSplat~\cite{in2024editsplat}. 
GaussianEditor employs the iterative dataset update strategy introduced in \cite{instructnerf2nerf}, while DGE, GaussCtrl and EditSplat adopt a joint multi-view image editing approach similar to our method.

\subsection{Qualitative Results}
We present the qualitative comparison in Fig.~\ref{fig:comp2} and Fig.~\ref{fig:comp}, showcasing editing results from two viewpoints. Compared to state-of-the-art methods~\cite{gaussianeditor2,dge,gaussctrl,in2024editsplat}, CoreEditor achieves better performance in producing vivid 3D edits that closely adhere to the text prompts. 
This advancement is largely due to the multi-view consistent 2D edits.

Based on the results, we can conclude that the inconsistency issues of other methods are mainly reflected in two aspects: 
(1) \textbf{Incomplete Editing:} When the edited training images are highly inconsistent, the resulting 3d scenes often exhibit insufficient visual change. 
For instance, GaussianEditor and EditSplat fail to modify the original GS model in the ``decayed skeleton horse'' cases in Fig.~\ref{fig:comp}. GaussCtrl exhibits similar limitations, as evidenced by its inability to successfully transform the human subject into an ape-man in Fig.~\ref{fig:comp2}.
(2) \textbf{Degraded Rendering Quality:} 
In some scenarios, although existing methods can partially achieve the target edits, they still produce locally inconsistent multi-view edits, resulting in blurry renderings with noticeable artifacts (highlighted by yellow dashed boxes in Fig.~\ref{fig:comp2} and Fig.~\ref{fig:comp}).
Specifically, for the stylization of $360^\circ$ scenes, such as the ``snow'' and ``Ukiyo-e'' cases in Fig.~\ref{fig:comp2}, the inconsistency introduces foggy artifacts that significantly degrade visual quality. Similarly, in the ``panda'' case in Fig.~\ref{fig:comp}, all the competitors produce vague panda faces.
In contrast, CoreEditor, equipped with the proposed CAA module and selective editing pipeline, effectively aligns multi-view edits at both global and local levels, significantly outperforming existing methods.
Besides, we also present a qualitative comparison with GaussCtrl~\cite{gaussctrl} using a free-viewpoint rendering video in the supplemental material. The comparison highlights that CoreEditor produces more faithful edits while substantially mitigating flickering artifacts. This further underscores the superior consistency brought by our method.

\begin{figure}[h]
    \centering
    \includegraphics[width=1\linewidth]{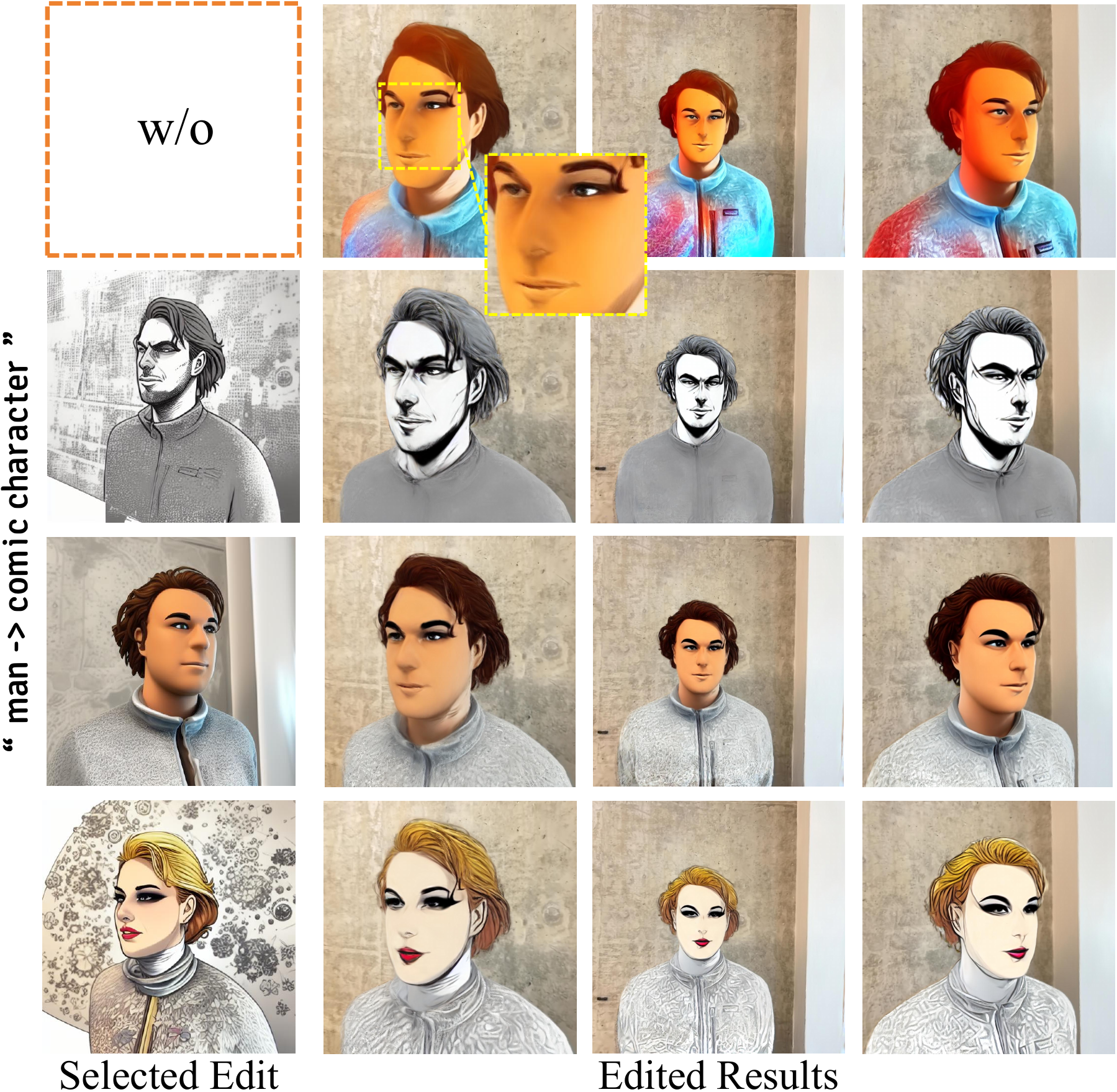}
    \caption{Results with different per-view edits $I^r$ selected.}
    \label{fig:ablaselection}
\end{figure}

\begin{figure*}[h]
  \centering
  \includegraphics[width=0.92\textwidth]{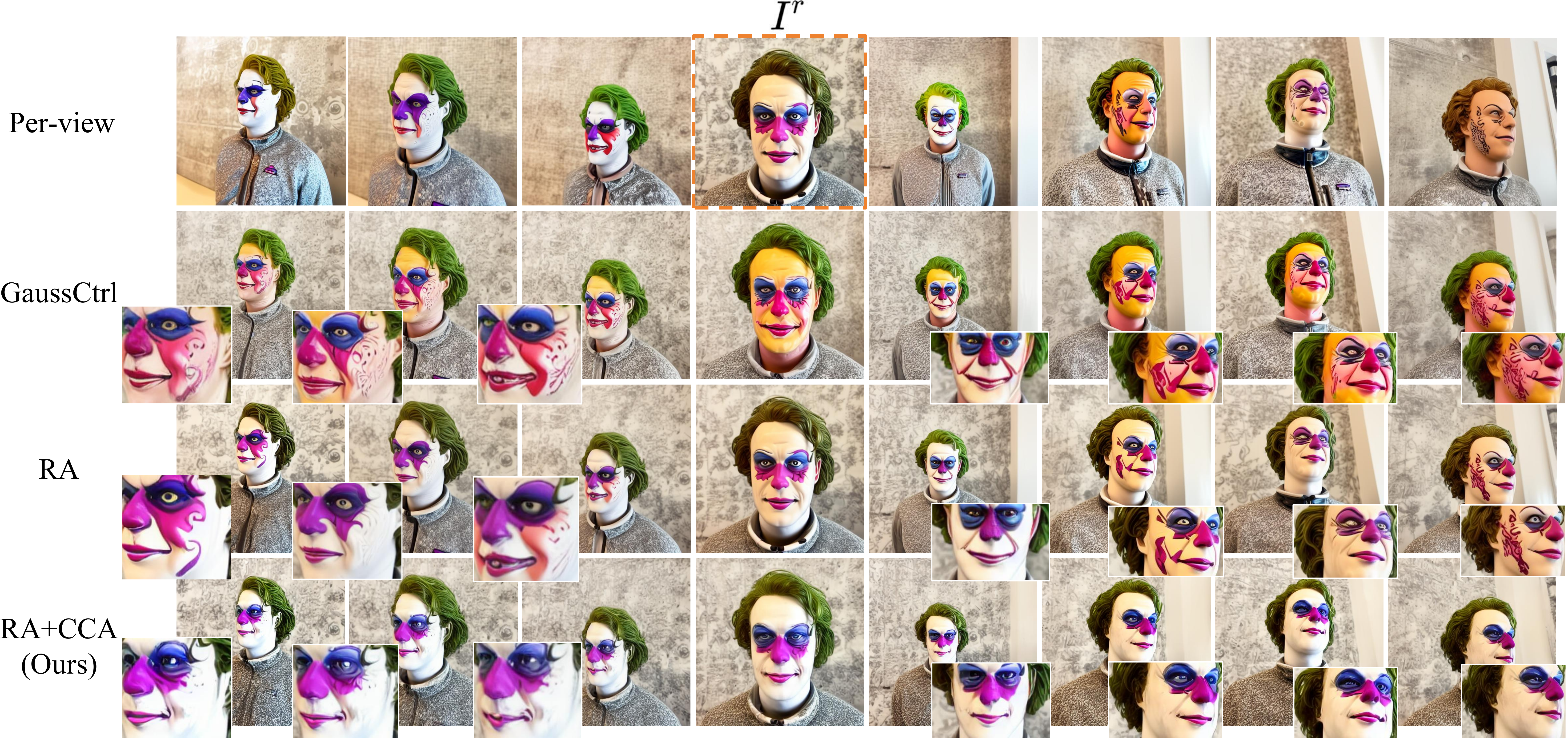}
\caption{Edited training images of \cite{gaussctrl} and different ablation variants for the ``joker'' prompt of the ``face'' scene. The selected edit $I^r$ is highlighted using a dash box.
Only CoreEditor ensures faithful propagation of the selected edit while maintaining 3D consistency across views.
}
  \label{fig:ablationCCA}
\end{figure*}

\begin{figure*}[h]
    \centering
    \includegraphics[width=0.9\linewidth]{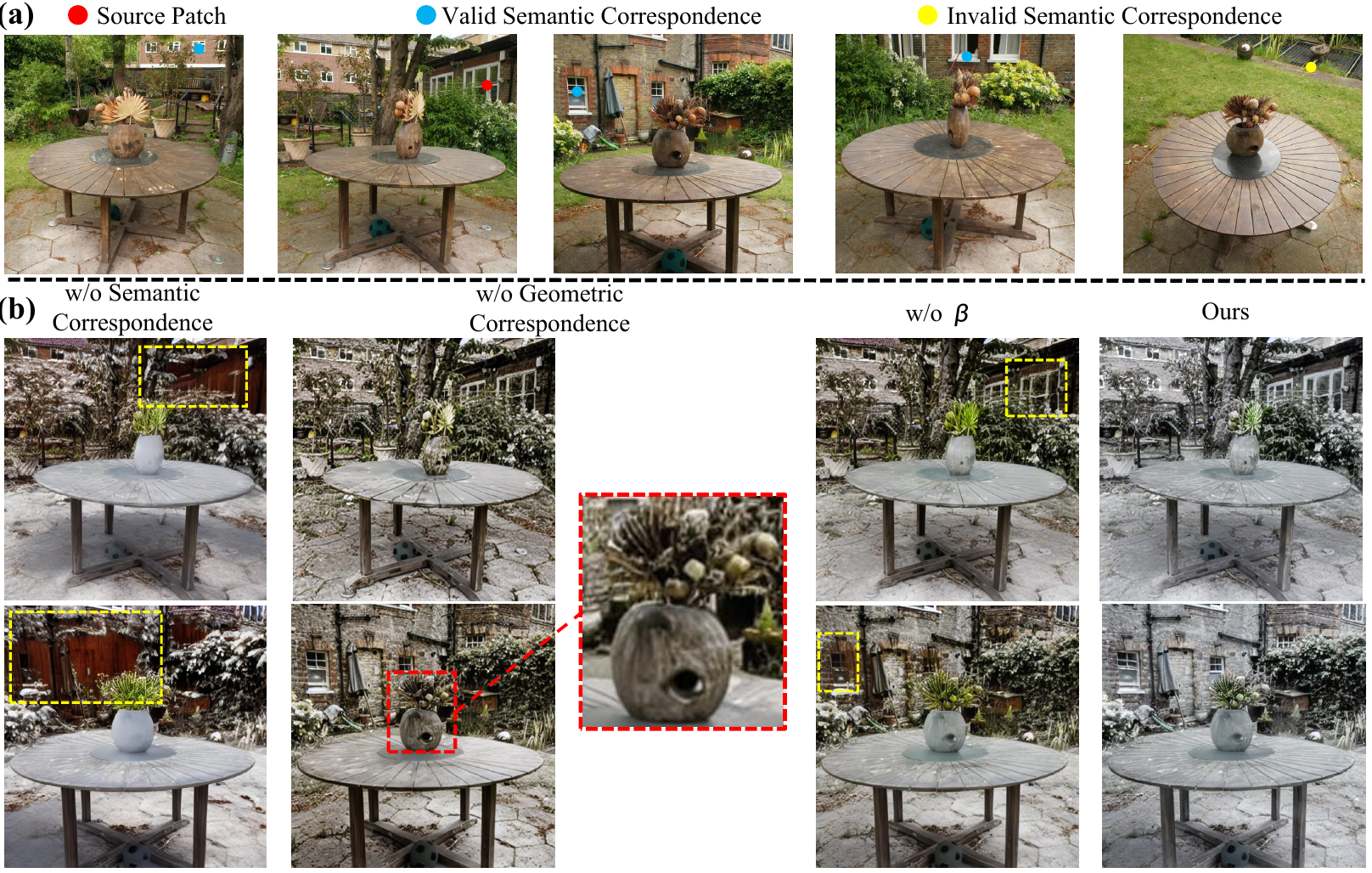}
    \caption{Ablation study of the co-supported correspondence. (a) Semantic correspondences obtained through the diffusion feature for the ``garden'' scene. Invalid correspondences filtered out by $\beta$ are marked as yellow. (b) Edited training images of different ablation variants. 
Without the semantic correspondence, the output images become unnatural and fail to preserve the original layout structure due to the insufficient attention token count.
}
    \label{fig:ablationSemantic}
\end{figure*}

\begin{figure*}[h]
    \centering
    \includegraphics[width=0.85\linewidth]{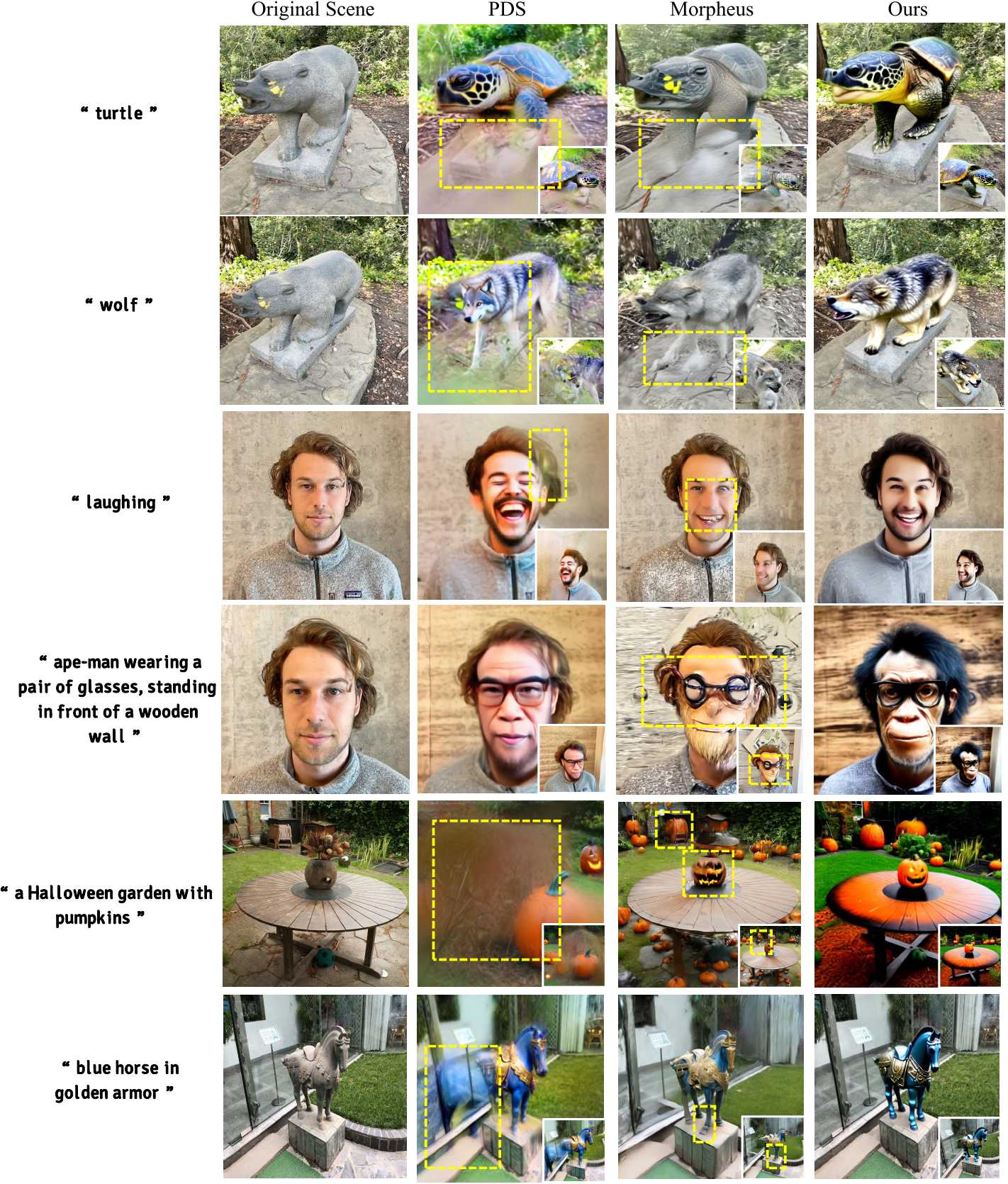}
    \caption{Visual comparison with PDS~\cite{pds} and Morpheus~\cite{wynn2025morpheus}.  We provide results rendered from two views for each edited scene. Blurry regions are highlighted with yellow dash boxes.
}
    \label{fig:compNew}
\end{figure*}

\begin{figure}[h]
  \centering
  \includegraphics[width=0.5\textwidth]{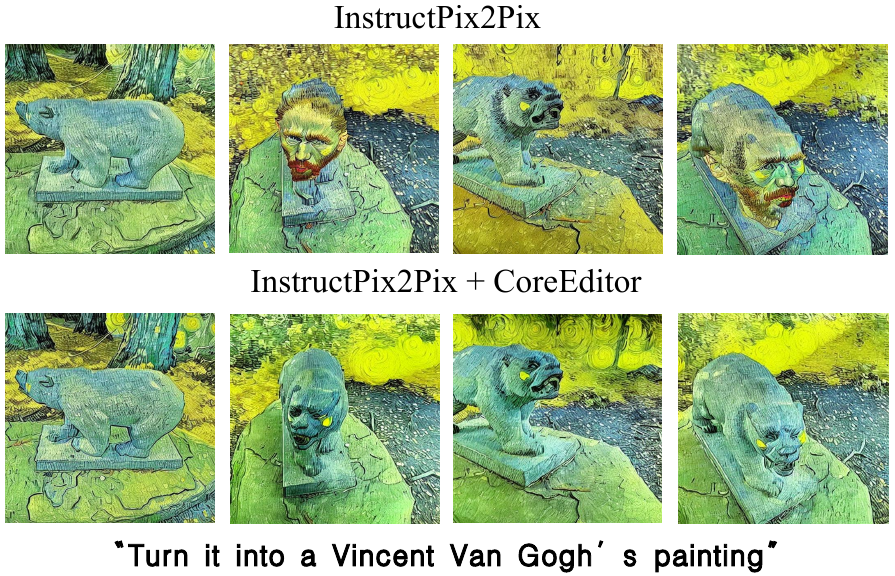}
\caption{Multi-view editing results produced by InstructPix2Pix~\cite{instructpix2pix} and its variant augmented with the proposed CoreEditor, which improves multi-view consistency.}
  \label{fig:ip2p}
\end{figure}

\begin{figure}[h]
  \centering
  \includegraphics[width=0.5\textwidth]{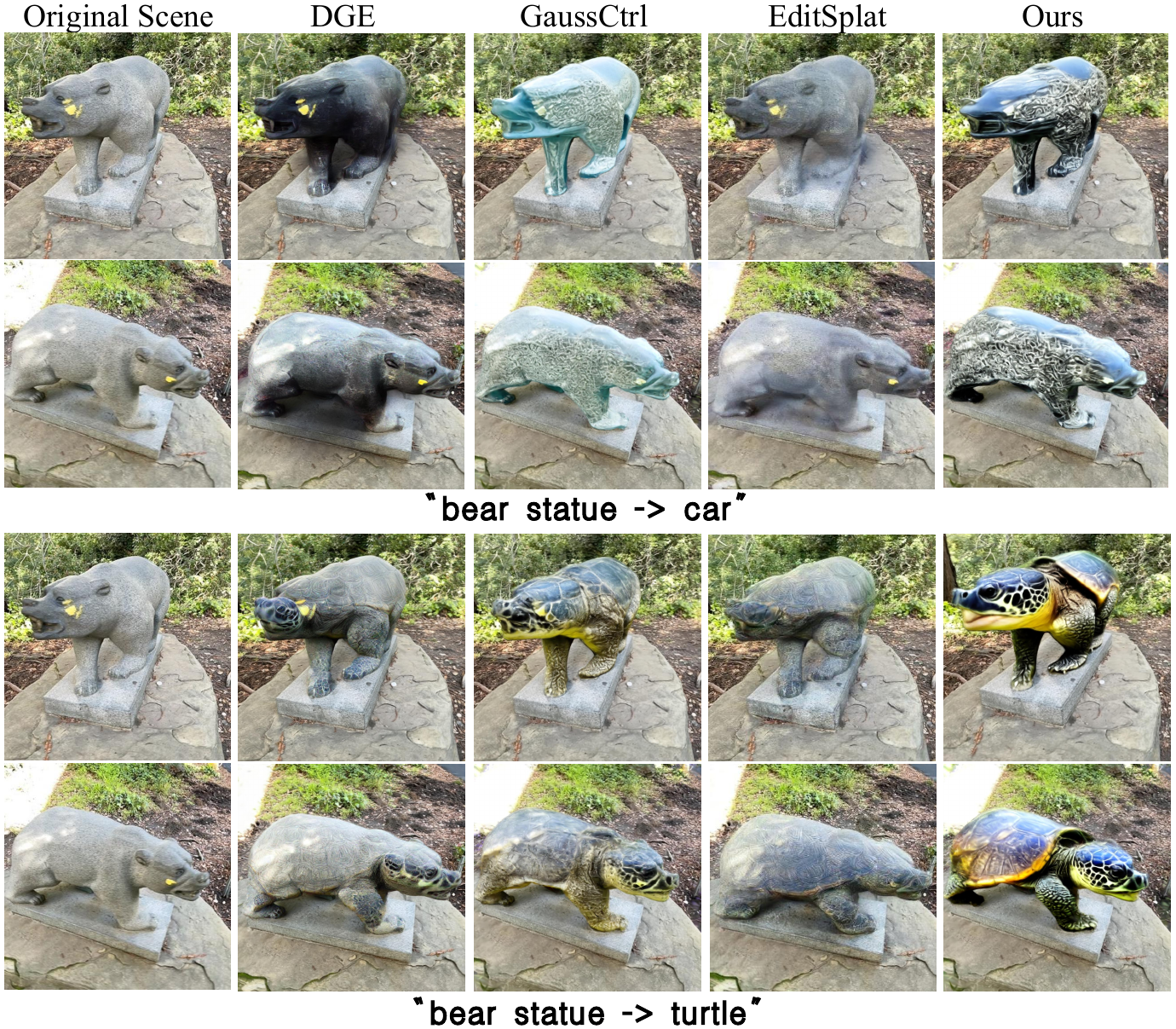}
\caption{Qualitative comparison with recent methods~\cite{dge,gaussctrl,in2024editsplat} on prompts requiring geometric changes.}
  \label{fig:case}
\end{figure}

\begin{figure*}[h]
  \centering
  \includegraphics[width=0.85\textwidth]{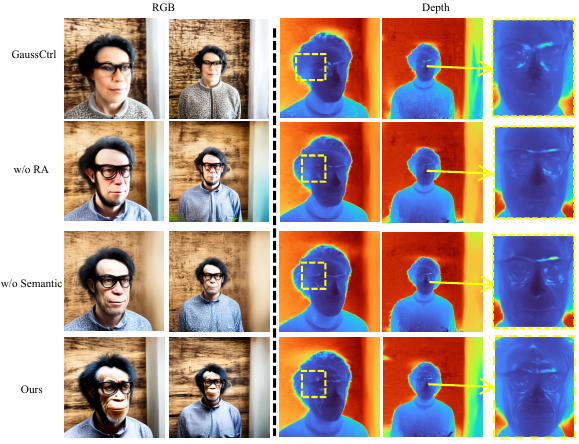}
\caption{Effect of each component on editing cases that require geometric changes. We show the rendered RGB and depth images for each view.}
  \label{fig:ablageo}
\end{figure*}

\subsection{Quantitative Results}
The quantitative comparison with baseline methods is summarized in Tab.~\ref{tab:comp}. Following the previous practice~\cite{instructnerf2nerf,gaussctrl,dge}, we evaluate the performance using two CLIP-based metrics computed on rendered images: the CLIP similarity score and the CLIP directional similarity score~\cite{instructnerf2nerf}. 
For each scene and prompt, CLIP scores are first averaged across rendered views to obtain a single score. We then report both the mean and the median across all evaluation cases, as shown in Tab.~\ref{tab:comp}.
The CLIP similarity score measures the degree of alignment between the edited images and the target text prompt, while the CLIP directional similarity score assesses how well the visual changes correspond to the semantic changes implied by the text.
As shown in Tab.~\ref{tab:comp}, CoreEditor consistently outperforms all baseline methods across both metrics, highlighting its superior ability to generate edits that are semantically faithful to the text prompt. 
To quantitatively evaluate the 3D consistency of CoreEditor, we employ the recently proposed Met3R metric~\cite{asim2025met3r}, which measures feature similarity between view-warped DINO~\cite{oquab2024dinov2} features.
We compute the Met3R values on the edited multi-view training images generated by DGE, GaussCtrl, EditSplat and our method. 
GaussianEditor is excluded from this comparison as it does not support joint multi-view image editing. 
The results demonstrate that CoreEditor significantly improves 3D consistency compared to baseline methods through the proposed techniques.
Moreover, given the inherently subjective nature of 3D editing, we validate our method through a user study involving 50 participants and 10 editing prompts. 
Given the source scene, editing prompts, and the rendered videos of each method, participants were asked to select the best method based on two criteria, respectively: (1) Overall visual quality and (2) 3D consistency (mainly based on the frequency of flickering artifacts in rendered videos). The user voting rates in Tab.~\ref{tab:comp} indicate a clear preference for CoreEditor’s outputs, further corroborating the advantages of our approach from a human perceptual perspective.

\noindent\textbf{Efficiency Comparison.}
Since the diffusion model in CoreEditor operates in a zero-shot manner, our multi-view joint editing only requires about 18 GB of GPU memory under the default settings. In terms of runtime, CoreEditor typically completes 3D scene editing within 8 minutes. While DGE achieves faster processing (5 minutes), other methods are slower than CoreEditor: GaussCtrl (10 minutes), EditSplat (12 minutes), and GaussianEditor (25 minutes). Despite not being the fastest, CoreEditor delivers significant improvements in editing quality while maintaining a computationally efficient design, avoiding additional overhead.

\subsection{Ablation Study}
\label{sec:abla}
We ablate the main components of CoreEditor by presenting quantitative and qualitative comparisons.

\noindent\textbf{Effect of Selective Editing Pipeline.}
In Fig.~\ref{fig:ablaselection}, we first evaluate the impact of selective editing by removing the selection process and disabling the RA module. 
As shown in the first row, rendering results exhibit unnatural color distribution when the selection stage is omitted. 
This issue arises because the per-view edits for the "comic character" prompt exhibit significant variations. Without selective editing, relying solely on a multi-view constraint like CAA leads to the blending of these highly varied edits, ultimately resulting in unnatural colors.
The decreased CLIP metrics in Tab.~\ref{tab:abla1} also indicate misaligned edits with the text prompts when the selective editing is removed.
To address this, we pre-align the global editing styles with the selected reference edit. As demonstrated in the subsequent rows of Fig.~\ref{fig:ablaselection}, selecting different $I^r$ enables CoreEditor to produce 3D edits with entirely distinct visual patterns. This flexibility facilitates a highly adaptable and user-centric editing process, empowering users to tailor edits more effectively to their desired visual styles or specific preferences.

\noindent\textbf{Manual vs. Automatic Selection.} 
While the selective editing design provides a user-centered editing experience, it introduces an additional manual step that may reduce workflow efficiency. To enable fully automatic editing within our CoreEditor, we replace manual selection with the ImageReward model~\cite{xu2023imagereward}, which evaluates generated images based on their alignment with the text prompt. Specifically, for a given target prompt, we rank all per-view edits using ImageReward and select the top-ranked edit as the reference.
The quantitative results in Tab.~\ref{tab:abla1} show that the performance of CoreEditor remains comparable regardless of whether manual or automatic selection is used. 
This robustness confirms that our selective editing framework flexibly accommodates both manual and automatic selection modes.
Furthermore, to ensure a comprehensive comparison with state-of-the-art methods, we integrate the RA module (with both manual and automatic selection) into GaussCtrl~\cite{gaussctrl} in Tab.~\ref{tab:abla1}. While the selective editing strategy improves GaussCtrl's performance, it still exhibits limitations in generating text-faithful edits. 
This is clearly demonstrated by its inferior CLIP scores and higher Met3R values compared to our approach, indicating that without consistency provided by our CCA, GaussCtrl cannot achieve edits that are faithfully aligned with prompts.

\begin{table}
        \renewcommand\arraystretch{1.2}
        \centering
        \caption{Effect of the selective editing pipeline. For CLIP-based metrics, we report both the mean and median values across all evaluation cases (shown as mean / median).}
        \label{tab:abla1}
        \small
        \begin{tabular}{|c|c|c| c |}
        \hline
        Methods  & \makebox[0.15\linewidth][c]{CLIP$_{sim}\uparrow$} & \makebox[0.15\linewidth][c]{CLIP$_{dir}\uparrow$} & \makebox[0.1\linewidth][c]{Met3R$\downarrow$}\\
        \hline
        GaussCtrl~\cite{gaussctrl}  & 0.257 / 0.251 & 0.128 / 0.126 &  0.372 \\
        \cite{gaussctrl} + RA (Automatic)  & 0.258 / 0.245 & 0.129 / 0.128 &  0.357 \\
        \cite{gaussctrl} + RA (Manual)  & 0.260 / 0.253 & 0.132 / 0.130 &  0.352 \\
        \hline
        Ours w/o RA  & 0.258 / 0.246 & 0.133 / 0.129 &  0.292 \\
        Ours (Automatic)  & 0.267 / 0.258 & \textbf{0.145} / 0.138 &  0.283 \\
        Ours (Manual) & \textbf{0.270} / \textbf{0.259} & \textbf{0.145} / \textbf{0.141} & \textbf{0.281} \\
        \hline
        \end{tabular}
\end{table}

\noindent\textbf{Effect of CCA.}
In Tab.~\ref{tab:abla2}, the Met3R value significantly increases when CCA is disabled, indicating the importance of CCA for ensuring multi-view consistency.
In Fig.~\ref{fig:ablationCCA}, we evaluate the effectiveness of the CCA module by comparing edited training images produced by different variants.
In the first row, the standard DDIM inversion-based editor is applied to edit multi-view images independently, view by view. Without any multi-view fusion strategy, the per-view edits exhibit significant inconsistency.
Next, as shown in the third row, we introduce the RA module to the basic 2D editor by selecting a reference edit $I^r$ (highlighted with a dash box). However, in the absence of a precise multi-view constraint, the RA variant is only capable of aligning visual patterns at a global level, failing to maintain consistency in local image details.
Finally, we incorporate the proposed CCA module into the diffusion model. With this precise multi-view constraint, the diffusion model effectively links edits across views, producing results with high-level consistency in both global patterns and local details, as demonstrated in the last row.

\begin{table}
        \renewcommand\arraystretch{1.2}
        \centering
        \caption{Effect of the Correspondence-constraint Attention. For CLIP-based metrics, we report both the mean and median values across all evaluation cases (shown as mean / median).}
        \label{tab:abla2}
        \small
        \begin{tabular}{|c|c|c| c |}
        \hline
        Methods  & \makebox[0.15\linewidth][c]{CLIP$_{sim}\uparrow$} & \makebox[0.15\linewidth][c]{CLIP$_{dir}\uparrow$} & \makebox[0.1\linewidth][c]{Met3R$\downarrow$}\\
        \hline
        w/o CAA & 0.250 / 0.247 & 0.124 / 0.126 &  0.378  \\
        w/o Geometric & 0.245 / 0.240 & 0.118 / 0.113 & 0.351 \\
        w/o Semantic & 0.266 / 0.255 & 0.128 / 0.129 & 0.294 \\
        Ours & \textbf{0.270} / \textbf{0.259} & \textbf{0.145} / \textbf{0.141} & \textbf{0.281} \\
        \hline
        \end{tabular}
\end{table}

\begin{table}
        \renewcommand\arraystretch{1.2}
        \centering
        \caption{Effect of varying $\beta$. For CLIP-based metrics, we report both the mean and median values across all evaluation cases (shown as mean / median).}
        \label{tab:ablabb}
        \small
        \begin{tabular}{|c|c|c| c |}
        \hline
        Methods  & \makebox[0.15\linewidth][c]{CLIP$_{sim}\uparrow$} & \makebox[0.15\linewidth][c]{CLIP$_{dir}\uparrow$} & \makebox[0.1\linewidth][c]{Met3R$\downarrow$}\\
        \hline
        w/o $\beta$ & 0.257 / 0.252 & 0.133 / 0.136 &  0.297  \\
        $\beta=0.8$ & 0.265 / 0.257 & 0.136 / 0.140 & 0.292 \\
        $\beta=0.85$ & \textbf{0.271} / 0.257 & 0.142 / 0.139 & 0.289 \\
        $\beta=0.9$ & 0.270 / \textbf{0.259} & \textbf{0.145} / \textbf{0.141} & \textbf{0.281} \\
        $\beta=0.95$ & 0.266 / 0.255 & 0.137 / 0.138 & 0.290 \\
        \hline
        \end{tabular}
\end{table}

\noindent\textbf{Effect of Co-supported Correspondence.}
We evaluate the co-supported correspondence by analyzing the 2D editing results of different variants, as shown in Fig.~\ref{fig:ablationSemantic}. 
We first visualize the semantic correspondences derived from diffusion features, which can accurately capture corresponding patches with similar semantic meanings, as demonstrated in Fig.~\ref{fig:ablationSemantic} (a).
Then, we assess the impact of removing semantic correspondences while relying solely on geometric information. In $360^\circ$ scenes like the ``garden'', geometric correspondences for background pixels are sparse due to the training data focusing on central objects. As illustrated in the first column of Fig.~\ref{fig:ablationSemantic} (b), the lack of tokens destabilizes the attention mechanism, resulting in distorted background and over-saturated colors.
This degraded visual quality can also be reflected by the decreased CLIP metrics in Tab.~\ref{tab:abla2}.
Next, we replace all geometric correspondences with semantic ones. As shown in the second column of Fig.~\ref{fig:ablationSemantic} (b) and the increased Met3R value in Tab.~\ref{tab:abla2}, this substitution leads to incomplete and inconsistent edits, as the well-reconstructed geometric correspondences are more accurate for central objects. 

\noindent\textbf{Effect of $\beta$.}
We firstly analyze the effect of the threshold $\beta$ by disabling it. Fig.\ref{fig:ablationSemantic} (a) indicates that some views lack patches with highly similar semantic meanings, and including such tokens may introduce noisy information, as evidenced by the excessive noise in the fourth column of Fig.\ref{fig:ablationSemantic} (b). Results in the first row of Tab.~\ref{tab:abla2} further quantify the effect of $\beta$.
We also ablate the value of $\beta$ in Tab.~\ref{tab:abla2} and find that CoreEditor achieves the best performance when $\beta=0.9$.

\begin{table}
        \renewcommand\arraystretch{1.2}
        \centering
        \caption{Additional quantitative comparison with PDS~\cite{pds} and Morpheus~\cite{wynn2025morpheus}. For CLIP-based metrics, we report both the mean and median values across all evaluation cases (shown as mean / median).}
        \label{tab:compNew}
        \small
        \begin{tabular}{|c|c|c| c |}
        \hline
        Methods  & \makebox[0.15\linewidth][c]{CLIP$_{sim}\uparrow$} & \makebox[0.15\linewidth][c]{CLIP$_{dir}\uparrow$} & \makebox[0.1\linewidth][c]{Met3R$\downarrow$}\\
        \hline
        PDS~\cite{pds} & 0.247 / 0.220 & 0.092 / 0.088 & - \\
        Morpheus~\cite{wynn2025morpheus} & 0.259 / 0.242 & 0.123 / 0.137 & 0.328 \\
        Ours & \textbf{0.270} / \textbf{0.259} & \textbf{0.145} / \textbf{0.141} & \textbf{0.281} \\
        \hline
        \end{tabular}
\end{table}

\subsection{Additional Comparisons}
In the main experiments, we compare CoreEditor with baselines that also apply the multi-view editing paradigm in a zero-shot manner. To more comprehensively evaluate our method, we include an SDS-based method~\cite{pds} and Morpheus~\cite{wynn2025morpheus} as baselines in Tab.~\ref{tab:compNew} and Fig.~\ref{fig:compNew}.

SDS-based methods often suffer from slow convergence and unstable optimization, resulting in noticeably lower rendering quality and significantly longer optimization time in practice. In contrast, Morpheus adopts a fundamentally different paradigm by finetuning a 2D diffusion model with RGB-D supervision to enhance geometry-aware editing. Although this strategy can induce more pronounced geometric changes than other baselines~\cite{gaussctrl,dge}, Morpheus directly edits depth through the diffusion process, which introduces strong geometric perturbations and often leads to degenerate 3D editing results. In addition, it finetunes a text-to-image diffusion model on relatively small-scale data, which disrupts the learned generative priors and leads to unstable editing behavior. Compared with these approaches, CoreEditor achieves a more favorable balance between editing strength, rendering quality, and robustness, delivering consistent high-quality results.

\subsection{Generalization to Different 2D Editors}

To demonstrate that the proposed CCA can generalize to different 2D editors, we integrate it into another commonly used 2D editor, InstructPix2Pix~\cite{instructpix2pix}. As shown in Fig.~\ref{fig:ip2p}, incorporating CCA leads to improved multi-view consistency in the outputs.

\subsection{Failure Cases and Limitations}
Similar to recent methods~\cite{gaussctrl,vica,rojas2025datenerf,in2024editsplat}, CoreEditor leverages the original scene geometry to ensure consistent 3D editing. As a result, our approach is limited in its ability to significantly modify scene geometry. As illustrated by the failure case in the first row of Fig.~\ref{fig:case}, both CoreEditor and other methods fail to transform the bear statue into a car. However, the second row of Fig.~\ref{fig:case} demonstrates that, for prompts requiring shape modification, our method surpasses baseline methods~\cite{dge,gaussctrl,in2024editsplat} by enabling greater geometric changes while maintaining high-quality rendering, thanks to our consistent 2D edits.

The stronger capability of CoreEditor to induce geometric changes compared to prior methods can be attributed to three key factors.
(1) Unlike methods such as~\cite{vica,rojas2025datenerf}, which directly warp pixels based on depth maps, CoreEditor enforces correspondence-based constraints within the diffusion latent space, where the minimal patch size is $8\times8$. This design makes CoreEditor more robust to depth inaccuracies and enables it to handle target edits whose geometric shapes deviate from the original input.
(2) Prompts that require geometric editing are often highly ambiguous, leading different views to produce diverse and sometimes conflicting editing results when processed independently. Existing multi-view editing methods~\cite{gaussctrl} tend to implicitly average these inconsistent outcomes across views, which suppresses distinct geometric changes and results in conservative edits. In contrast, our selective editing pipeline injects the most representative editing mode through the RA module, avoiding undesired averaging effects and facilitating more pronounced and coherent geometric transformations.
(3) In addition, the proposed semantic correspondence allows CCA to better associate semantically similar regions across different views, enabling more consistent feature alignment during editing. By explicitly linking corresponding semantic regions, CCA effectively propagates geometry-related changes across views, leading to more stable and coherent geometric editing results.

As further evidence, Fig.~\ref{fig:ablageo} compares CoreEditor with GaussCtrl~\cite{gaussctrl} and our ablation variants on geometric editing, showing that both RA and semantic correspondence are crucial for inducing geometric deformations. Removing either component noticeably weakens the extent of geometric changes, while their combination enables CoreEditor to achieve stronger geometric editing results than existing methods.

\section{Conclusion}
We propose CoreEditor, a novel framework for text-driven 3D editing that significantly enhances the quality of edited 3D scenes by improving consistency during multi-view editing. At the heart of our method is the correspondence-constrained attention module, which enforces interactions between image patches that should remain consistent within the diffusion model. To handle complex scenes with the proposed attention module, we introduce a geometric and semantic co-supported strategy to extract comprehensive correspondences, ensuring robust multi-view editing.
Additionally, we design a selective editing pipeline that empowers users to choose their preferred edits from multiple candidates, enabling a highly flexible and user-centric editing process. Extensive experiments on widely-used datasets demonstrate that CoreEditor achieves state-of-the-art editing performance, offering superior quality and adaptability compared to existing methods.
While the current implementation of CCA is designed for U-Net–based 2D diffusion editors, extending it to architectures that incorporate explicit positional encoding within the attention mechanism (e.g., FLUX~\cite{flux2024}) would require non-trivial architectural adaptations to jointly account for positional information and cross-view correspondence. We leave this direction for future work.

\section{Acknowledgments}
This work was supported by the National Natural Science Foundation of China (No. T2322012, No. 62572240, 62172218), and the Shenzhen Science and Technology Program
(No. JCYJ20220818103401003, No. JCYJ20220530172403007).




\bibliographystyle{IEEEtran}
\bibliography{main}

@article{nerf,
  title={Nerf: Representing scenes as neural radiance fields for view synthesis},
  author={Mildenhall, Ben and Srinivasan, Pratul P and Tancik, Matthew and Barron, Jonathan T and Ramamoorthi, Ravi and Ng, Ren},
  journal={Communications of the ACM},
  volume={65},
  number={1},
  pages={99--106},
  year={2021}
}

@inproceedings{asim2025met3r,
  title={Met3r: Measuring multi-view consistency in generated images},
  author={Asim, Mohammad and Wewer, Christopher and Wimmer, Thomas and Schiele, Bernt and Lenssen, Jan Eric},
  booktitle={Proceedings of the IEEE/CVF Computer Vision and Pattern Recognition Conference},
  pages={6034--6044},
  year={2025}
}

@inproceedings{wynn2025morpheus,
  title={Morpheus: Text-Driven 3D Gaussian Splat Shape and Color Stylization},
  author={Wynn, Jamie and Qureshi, Zawar and Powierza, Jakub and Watson, Jamie and Sayed, Mohamed},
  booktitle={Proceedings of the IEEE/CVF Computer Vision and Pattern Recognition Conference},
  pages={7825--7836},
  year={2025}
}

@inproceedings{pds,
  title={Posterior distillation sampling},
  author={Koo, Juil and Park, Chanho and Sung, Minhyuk},
  booktitle={Proceedings of the IEEE/CVF Conference on Computer Vision and Pattern Recognition},
  pages={13352--13361},
  year={2024}
}

@inproceedings{dds,
  title={Delta denoising score},
  author={Hertz, Amir and Aberman, Kfir and Cohen-Or, Daniel},
  booktitle={Proceedings of the IEEE/CVF International Conference on Computer Vision},
  pages={2328--2337},
  year={2023}
}

@article{3dgs,
  title={3d gaussian splatting for real-time radiance field rendering.},
  author={Kerbl, Bernhard and Kopanas, Georgios and Leimk{\"u}hler, Thomas and Drettakis, George},
  journal={ACM Trans. Graph.},
  volume={42},
  number={4},
  pages={139--1},
  year={2023}
}

@inproceedings{latentdiffusion,
  title={High-resolution image synthesis with latent diffusion models},
  author={Rombach, Robin and Blattmann, Andreas and Lorenz, Dominik and Esser, Patrick and Ommer, Bj{\"o}rn},
  booktitle={Proceedings of the IEEE/CVF conference on computer vision and pattern recognition},
  pages={10684--10695},
  year={2022}
}

@inproceedings{controlnet,
  title={Adding conditional control to text-to-image diffusion models},
  author={Zhang, Lvmin and Rao, Anyi and Agrawala, Maneesh},
  booktitle={Proceedings of the IEEE/CVF International Conference on Computer Vision},
  pages={3836--3847},
  year={2023}
}

@inproceedings{instructpix2pix,
  title={Instructpix2pix: Learning to follow image editing instructions},
  author={Brooks, Tim and Holynski, Aleksander and Efros, Alexei A},
  booktitle={Proceedings of the IEEE/CVF Conference on Computer Vision and Pattern Recognition},
  pages={18392--18402},
  year={2023}
}

@inproceedings{instructnerf2nerf,
  title={Instruct-nerf2nerf: Editing 3d scenes with instructions},
  author={Haque, Ayaan and Tancik, Matthew and Efros, Alexei A and Holynski, Aleksander and Kanazawa, Angjoo},
  booktitle={Proceedings of the IEEE/CVF International Conference on Computer Vision},
  pages={19740--19750},
  year={2023}
}

@article{vica,
  title={Vica-nerf: View-consistency-aware 3d editing of neural radiance fields},
  author={Dong, Jiahua and Wang, Yu-Xiong},
  journal={Advances in Neural Information Processing Systems},
  volume={36},
  year={2024}
}

@inproceedings{gaussctrl,
  title={Gaussctrl: Multi-view consistent text-driven 3d gaussian splatting editing},
  author={Wu, Jing and Bian, Jia-Wang and Li, Xinghui and Wang, Guangrun and Reid, Ian and Torr, Philip and Prisacariu, Victor Adrian},
  booktitle={European Conference on Computer Vision},
  pages={55--71},
  year={2024},
  organization={Springer}
}

@inproceedings{dge,
  title={Dge: Direct gaussian 3d editing by consistent multi-view editing},
  author={Chen, Minghao and Laina, Iro and Vedaldi, Andrea},
  booktitle={European Conference on Computer Vision},
  pages={74--92},
  year={2024},
  organization={Springer}
}

@inproceedings{rojas2025datenerf,
  title={Datenerf: Depth-aware text-based editing of nerfs},
  author={Rojas, Sara and Philip, Julien and Zhang, Kai and Bi, Sai and Luan, Fujun and Ghanem, Bernard and Sunkavalli, Kalyan},
  booktitle={European Conference on Computer Vision},
  pages={267--284},
  year={2024},
  organization={Springer}
}

@inproceedings{chen2024consistdreamer,
  title={ConsistDreamer: 3D-Consistent 2D Diffusion for High-Fidelity Scene Editing},
  author={Chen, Jun-Kun and Bul{\`o}, Samuel Rota and M{\"u}ller, Norman and Porzi, Lorenzo and Kontschieder, Peter and Wang, Yu-Xiong},
  booktitle={Proceedings of the IEEE/CVF Conference on Computer Vision and Pattern Recognition},
  pages={21071--21080},
  year={2024}
}

@article{cai2024mv2mv,
  title={MV2MV: Multi-View Image Translation via View-Consistent Diffusion Models},
  author={Cai, Youcheng and Li, Runshi and Liu, Ligang},
  journal={ACM Transactions on Graphics},
  volume={43},
  number={6},
  pages={1--12},
  year={2024},
  publisher={ACM New York, NY, USA}
}

@inproceedings{luo20253denhancer,
  title={3DEnhancer: Consistent Multi-View Diffusion for 3D Enhancement},
  author={Luo, Yihang and Zhou, Shangchen and Lan, Yushi and Pan, Xingang and Loy, Chen Change},
  booktitle={Proceedings of the Computer Vision and Pattern Recognition Conference},
  pages={16430--16440},
  year={2025}
}

@article{xu2023imagereward,
  title={Imagereward: Learning and evaluating human preferences for text-to-image generation},
  author={Xu, Jiazheng and Liu, Xiao and Wu, Yuchen and Tong, Yuxuan and Li, Qinkai and Ding, Ming and Tang, Jie and Dong, Yuxiao},
  journal={Advances in Neural Information Processing Systems},
  volume={36},
  pages={15903--15935},
  year={2023}
}

@inproceedings{tuneavideo,
  title={Tune-a-video: One-shot tuning of image diffusion models for text-to-video generation},
  author={Wu, Jay Zhangjie and Ge, Yixiao and Wang, Xintao and Lei, Stan Weixian and Gu, Yuchao and Shi, Yufei and Hsu, Wynne and Shan, Ying and Qie, Xiaohu and Shou, Mike Zheng},
  booktitle={Proceedings of the IEEE/CVF International Conference on Computer Vision},
  pages={7623--7633},
  year={2023}
}

@inproceedings{
    dift,
    title={Emergent Correspondence from Image Diffusion},
    author={Luming Tang and Menglin Jia and Qianqian Wang and Cheng Perng Phoo and Bharath Hariharan},
    booktitle={Thirty-seventh Conference on Neural Information Processing Systems},
    year={2023},
    url={https://openreview.net/forum?id=ypOiXjdfnU}
}

@inproceedings{clipnerf,
  title={Clip-nerf: Text-and-image driven manipulation of neural radiance fields},
  author={Wang, Can and Chai, Menglei and He, Mingming and Chen, Dongdong and Liao, Jing},
  booktitle={Proceedings of the IEEE/CVF Conference on Computer Vision and Pattern Recognition},
  pages={3835--3844},
  year={2022}
}

@ARTICLE{nerfart,
  author={Wang, Can and Jiang, Ruixiang and Chai, Menglei and He, Mingming and Chen, Dongdong and Liao, Jing},
  journal={IEEE Transactions on Visualization and Computer Graphics}, 
  title={NeRF-Art: Text-Driven Neural Radiance Fields Stylization}, 
  year={2024},
  volume={30},
  number={8},
  pages={4983-4996}
}

@inproceedings{text2mesh,
  title={Text2mesh: Text-driven neural stylization for meshes},
  author={Michel, Oscar and Bar-On, Roi and Liu, Richard and Benaim, Sagie and Hanocka, Rana},
  booktitle={Proceedings of the IEEE/CVF Conference on Computer Vision and Pattern Recognition},
  pages={13492--13502},
  year={2022}
}

@inproceedings{clip,
  title={Learning transferable visual models from natural language supervision},
  author={Radford, Alec and Kim, Jong Wook and Hallacy, Chris and Ramesh, Aditya and Goh, Gabriel and Agarwal, Sandhini and Sastry, Girish and Askell, Amanda and Mishkin, Pamela and Clark, Jack and others},
  booktitle={International conference on machine learning},
  pages={8748--8763},
  year={2021}
}

@inproceedings{
dreamfusion,
title={DreamFusion: Text-to-3D using 2D Diffusion},
author={Ben Poole and Ajay Jain and Jonathan T. Barron and Ben Mildenhall},
booktitle={The Eleventh International Conference on Learning Representations },
year={2023}
}

@inproceedings{voxe,
  title={Vox-e: Text-guided voxel editing of 3d objects},
  author={Sella, Etai and Fiebelman, Gal and Hedman, Peter and Averbuch-Elor, Hadar},
  booktitle={Proceedings of the IEEE/CVF International Conference on Computer Vision},
  pages={430--440},
  year={2023}
}

@inproceedings{dreameditor,
  title={Dreameditor: Text-driven 3d scene editing with neural fields},
  author={Zhuang, Jingyu and Wang, Chen and Lin, Liang and Liu, Lingjie and Li, Guanbin},
  booktitle={SIGGRAPH Asia 2023 Conference Papers},
  pages={1--10},
  year={2023}
}

@inproceedings{watchyourstep,
  title={Watch your steps: Local image and scene editing by text instructions},
  author={Mirzaei, Ashkan and Aumentado-Armstrong, Tristan and Brubaker, Marcus A and Kelly, Jonathan and Levinshtein, Alex and Derpanis, Konstantinos G and Gilitschenski, Igor},
  booktitle={European Conference on Computer Vision},
  pages={111--129},
  year={2024}
}

@article{tipeditor,
  title={Tip-editor: An accurate 3d editor following both text-prompts and image-prompts},
  author={Zhuang, Jingyu and Kang, Di and Cao, Yan-Pei and Li, Guanbin and Lin, Liang and Shan, Ying},
  journal={ACM Transactions on Graphics (TOG)},
  volume={43},
  number={4},
  pages={1--12},
  year={2024}
}

@inproceedings{gaussianeditor1,
  title={Gaussianeditor: Editing 3d gaussians delicately with text instructions},
  author={Wang, Junjie and Fang, Jiemin and Zhang, Xiaopeng and Xie, Lingxi and Tian, Qi},
  booktitle={Proceedings of the IEEE/CVF Conference on Computer Vision and Pattern Recognition},
  pages={20902--20911},
  year={2024}
}

@inproceedings{gaussianeditor2,
  title={Gaussianeditor: Swift and controllable 3d editing with gaussian splatting},
  author={Chen, Yiwen and Chen, Zilong and Zhang, Chi and Wang, Feng and Yang, Xiaofeng and Wang, Yikai and Cai, Zhongang and Yang, Lei and Liu, Huaping and Lin, Guosheng},
  booktitle={Proceedings of the IEEE/CVF Conference on Computer Vision and Pattern Recognition},
  pages={21476--21485},
  year={2024}
}

@inproceedings{vcedit,
  title={View-consistent 3d editing with gaussian splatting},
  author={Wang, Yuxuan and Yi, Xuanyu and Wu, Zike and Zhao, Na and Chen, Long and Zhang, Hanwang},
  booktitle={European Conference on Computer Vision},
  pages={404--420},
  year={2024}
}

@inproceedings{shapeditor,
  title={SHAP-EDITOR: Instruction-guided Latent 3D Editing in Seconds},
  author={Chen, Minghao and Xie, Junyu and Laina, Iro and Vedaldi, Andrea},
  booktitle={Proceedings of the IEEE/CVF Conference on Computer Vision and Pattern Recognition},
  pages={26456--26466},
  year={2024}
}

@inproceedings{freditor,
  title={Freditor: High-Fidelity and Transferable NeRF Editing by Frequency Decomposition},
  author={He, Yisheng and Yuan, Weihao and Zhu, Siyu and Dong, Zilong and Bo, Liefeng and Huang, Qixing},
  booktitle={European Conference on Computer Vision},
  pages={73--91},
  year={2024}
}

@inproceedings{latenteditor,
  title={LatentEditor: text driven local editing of 3D scenes},
  author={Khalid, Umar and Iqbal, Hasan and Karim, Nazmul and Tayyab, Muhammad and Hua, Jing and Chen, Chen},
  booktitle={European Conference on Computer Vision},
  pages={364--380},
  year={2024}
}

@inproceedings{zero123,
  title={Zero-1-to-3: Zero-shot one image to 3d object},
  author={Liu, Ruoshi and Wu, Rundi and Van Hoorick, Basile and Tokmakov, Pavel and Zakharov, Sergey and Vondrick, Carl},
  booktitle={Proceedings of the IEEE/CVF international conference on computer vision},
  pages={9298--9309},
  year={2023}
}

@inproceedings{
syncdreamer,
title={SyncDreamer: Generating Multiview-consistent Images from a Single-view Image},
author={Yuan Liu and Cheng Lin and Zijiao Zeng and Xiaoxiao Long and Lingjie Liu and Taku Komura and Wenping Wang},
booktitle={The Twelfth International Conference on Learning Representations},
year={2024}
}

@inproceedings{wen2025intergsedit,
  title={Intergsedit: Interactive 3d gaussian splatting editing with 3d geometry-consistent attention prior},
  author={Wen, Minghao and Wu, Shengjie and Wang, Kangkan and Liang, Dong},
  booktitle={Proceedings of the IEEE/CVF International Conference on Computer Vision},
  pages={26136--26145},
  year={2025}
}

@inproceedings{
shi2024mvdream,
title={{MVD}ream: Multi-view Diffusion for 3D Generation},
author={Yichun Shi and Peng Wang and Jianglong Ye and Long Mai and Kejie Li and Xiao Yang},
booktitle={The Twelfth International Conference on Learning Representations},
year={2024}
}

@misc{flux2024,
    author={Black Forest Labs},
    title={FLUX},
    year={2024},
    howpublished={\url{https://github.com/black-forest-labs/flux}},
}

@article{imagedream,
  title={Imagedream: Image-prompt multi-view diffusion for 3d generation},
  author={Wang, Peng and Shi, Yichun},
  journal={arXiv preprint arXiv:2312.02201},
  year={2023}
}

@inproceedings{
mvdiffusion,
title={{MVD}iffusion: Enabling Holistic Multi-view Image Generation with Correspondence-Aware Diffusion},
author={Shitao Tang and Fuyang Zhang and Jiacheng Chen and Peng Wang and Yasutaka Furukawa},
booktitle={Thirty-seventh Conference on Neural Information Processing Systems},
year={2023}
}

@inproceedings{spad,
  title={SPAD: Spatially Aware Multi-View Diffusers},
  author={Kant, Yash and Siarohin, Aliaksandr and Wu, Ziyi and Vasilkovsky, Michael and Qian, Guocheng and Ren, Jian and Guler, Riza Alp and Ghanem, Bernard and Tulyakov, Sergey and Gilitschenski, Igor},
  booktitle={Proceedings of the IEEE/CVF Conference on Computer Vision and Pattern Recognition},
  pages={10026--10038},
  year={2024}
}

@inproceedings{epidiff,
  title={Epidiff: Enhancing multi-view synthesis via localized epipolar-constrained diffusion},
  author={Huang, Zehuan and Wen, Hao and Dong, Junting and Wang, Yaohui and Li, Yangguang and Chen, Xinyuan and Cao, Yan-Pei and Liang, Ding and Qiao, Yu and Dai, Bo and others},
  booktitle={Proceedings of the IEEE/CVF Conference on Computer Vision and Pattern Recognition},
  pages={9784--9794},
  year={2024}
}

@inproceedings{eschernet,
  title={Eschernet: A generative model for scalable view synthesis},
  author={Kong, Xin and Liu, Shikun and Lyu, Xiaoyang and Taher, Marwan and Qi, Xiaojuan and Davison, Andrew J},
  booktitle={Proceedings of the IEEE/CVF Conference on Computer Vision and Pattern Recognition},
  pages={9503--9513},
  year={2024}
}

@article{ddpm,
  title={Denoising diffusion probabilistic models},
  author={Ho, Jonathan and Jain, Ajay and Abbeel, Pieter},
  journal={Advances in neural information processing systems},
  volume={33},
  pages={6840--6851},
  year={2020}
}

@inproceedings{latentdiff,
  title={High-resolution image synthesis with latent diffusion models},
  author={Rombach, Robin and Blattmann, Andreas and Lorenz, Dominik and Esser, Patrick and Ommer, Bj{\"o}rn},
  booktitle={Proceedings of the IEEE/CVF conference on computer vision and pattern recognition},
  pages={10684--10695},
  year={2022}
}

@inproceedings{unet,
  title={U-net: Convolutional networks for biomedical image segmentation},
  author={Ronneberger, Olaf and Fischer, Philipp and Brox, Thomas},
  booktitle={MICCAI},
  pages={234--241},
  year={2015}
}

@inproceedings{
ddim,
title={Denoising Diffusion Implicit Models},
author={Jiaming Song and Chenlin Meng and Stefano Ermon},
booktitle={International Conference on Learning Representations},
year={2021}
}

@inproceedings{pnp,
  title={Plug-and-play diffusion features for text-driven image-to-image translation},
  author={Tumanyan, Narek and Geyer, Michal and Bagon, Shai and Dekel, Tali},
  booktitle={Proceedings of the IEEE/CVF Conference on Computer Vision and Pattern Recognition},
  pages={1921--1930},
  year={2023}
}

@article{p2p,
  title={Prompt-to-prompt image editing with cross attention control},
  author={Hertz, Amir and Mokady, Ron and Tenenbaum, Jay and Aberman, Kfir and Pritch, Yael and Cohen-Or, Daniel},
  journal={arXiv preprint arXiv:2208.01626},
  year={2022}
}

@article{flatten,
  title={Flatten: optical flow-guided attention for consistent text-to-video editing},
  author={Cong, Yuren and Xu, Mengmeng and Simon, Christian and Chen, Shoufa and Ren, Jiawei and Xie, Yanping and Perez-Rua, Juan-Manuel and Rosenhahn, Bodo and Xiang, Tao and He, Sen},
  journal={arXiv preprint arXiv:2310.05922},
  year={2023}
}

@inproceedings{dreammatcher,
  title={Dreammatcher: Appearance matching self-attention for semantically-consistent text-to-image personalization},
  author={Nam, Jisu and Kim, Heesu and Lee, DongJae and Jin, Siyoon and Kim, Seungryong and Chang, Seunggyu},
  booktitle={Proceedings of the IEEE/CVF Conference on Computer Vision and Pattern Recognition},
  pages={8100--8110},
  year={2024}
}

@inproceedings{masactrl,
  title={Masactrl: Tuning-free mutual self-attention control for consistent image synthesis and editing},
  author={Cao, Mingdeng and Wang, Xintao and Qi, Zhongang and Shan, Ying and Qie, Xiaohu and Zheng, Yinqiang},
  booktitle={Proceedings of the IEEE/CVF International Conference on Computer Vision},
  pages={22560--22570},
  year={2023}
}

@article{Zhou2024storydiffusion,
  title={StoryDiffusion: Consistent Self-Attention for Long-Range Image and Video Generation},
  author={Zhou, Yupeng and Zhou, Daquan and Cheng, Ming-Ming and Feng, Jiashi and Hou, Qibin},
  year={2024}
}

@InProceedings{mipnerf360,
    author    = {Barron, Jonathan T. and Mildenhall, Ben and Verbin, Dor and Srinivasan, Pratul P. and Hedman, Peter},
    title     = {Mip-NeRF 360: Unbounded Anti-Aliased Neural Radiance Fields},
    booktitle = {Proceedings of the IEEE/CVF Conference on Computer Vision and Pattern Recognition},
    year      = {2022},
    pages     = {5470-5479}
}

@inproceedings{blendedmvs,
  title={Blendedmvs: A large-scale dataset for generalized multi-view stereo networks},
  author={Yao, Yao and Luo, Zixin and Li, Shiwei and Zhang, Jingyang and Ren, Yufan and Zhou, Lei and Fang, Tian and Quan, Long},
  booktitle={Proceedings of the IEEE/CVF conference on computer vision and pattern recognition},
  pages={1790--1799},
  year={2020}
}

@inproceedings{nerfstudio,
	title        = {Nerfstudio: A Modular Framework for Neural Radiance Field Development},
	author       = {
		Tancik, Matthew and Weber, Ethan and Ng, Evonne and Li, Ruilong and Yi, Brent
		and Kerr, Justin and Wang, Terrance and Kristoffersen, Alexander and Austin,
		Jake and Salahi, Kamyar and Ahuja, Abhik and McAllister, David and Kanazawa,
		Angjoo
	},
	year         = 2023,
	booktitle    = {ACM SIGGRAPH 2023 Conference Proceedings},
	series       = {SIGGRAPH '23}
}

@misc{diffusers,
  author = {Patrick von Platen and Suraj Patil and Anton Lozhkov and Pedro Cuenca and Nathan Lambert and Kashif Rasul and Mishig Davaadorj and Dhruv Nair and Sayak Paul and William Berman and Yiyi Xu and Steven Liu and Thomas Wolf},
  title = {Diffusers: State-of-the-art diffusion models},
  year = {2022},
  publisher = {GitHub},
  journal = {GitHub repository},
  howpublished = {\url{https://github.com/huggingface/diffusers}}
}

@inproceedings{SAM,
  title={Segment anything},
  author={Kirillov, Alexander and Mintun, Eric and Ravi, Nikhila and Mao, Hanzi and Rolland, Chloe and Gustafson, Laura and Xiao, Tete and Whitehead, Spencer and Berg, Alexander C and Lo, Wan-Yen and others},
  booktitle={Proceedings of the IEEE/CVF International Conference on Computer Vision},
  pages={4015--4026},
  year={2023}
}

@inproceedings{lpips,
  title={The unreasonable effectiveness of deep features as a perceptual metric},
  author={Zhang, Richard and Isola, Phillip and Efros, Alexei A and Shechtman, Eli and Wang, Oliver},
  booktitle={Proceedings of the IEEE conference on computer vision and pattern recognition},
  pages={586--595},
  year={2018}
}

@inproceedings{chen2024mvip,
  title={MVIP-NeRF: Multi-view 3D Inpainting on NeRF Scenes via Diffusion Prior},
  author={Chen, Honghua and Loy, Chen Change and Pan, Xingang},
  booktitle={Proceedings of the IEEE/CVF Conference on Computer Vision and Pattern Recognition},
  pages={5344--5353},
  year={2024}
}

@article{li2024era3d,
  title={Era3d: high-resolution multiview diffusion using efficient row-wise attention},
  author={Li, Peng and Liu, Yuan and Long, Xiaoxiao and Zhang, Feihu and Lin, Cheng and Li, Mengfei and Qi, Xingqun and Zhang, Shanghang and Xue, Wei and Luo, Wenhan and others},
  journal={Advances in Neural Information Processing Systems},
  volume={37},
  pages={55975--56000},
  year={2024}
}

@inproceedings{proedit,
 author = {Chen, Jun-Kun and Wang, Yu-Xiong},
 booktitle = {Advances in Neural Information Processing Systems},
 pages = {4934--4955},
 title = {ProEdit: Simple Progression is All You Need for High-Quality 3D Scene Editing},
 volume = {37},
 year = {2024}
}

@ARTICLE{gaussedit,
  author={Shu, Zhenyu and Yu, Junlong and Chao, Kai and Xin, Shiqing and Liu, Ligang},
  journal={IEEE Transactions on Visualization and Computer Graphics}, 
  title={GaussEdit: Adaptive 3D Scene Editing with Text and Image Prompts}, 
  year={2025},
  volume={},
  number={},
  pages={1-12}}

@article{huang2024mv,
  title={Mv-adapter: Multi-view consistent image generation made easy},
  author={Huang, Zehuan and Guo, Yuan-Chen and Wang, Haoran and Yi, Ran and Ma, Lizhuang and Cao, Yan-Pei and Sheng, Lu},
  journal={arXiv preprint arXiv:2412.03632},
  year={2024}
}

@inproceedings{he2024customize,
  title={Customize your nerf: Adaptive source driven 3d scene editing via local-global iterative training},
  author={He, Runze and Huang, Shaofei and Nie, Xuecheng and Hui, Tianrui and Liu, Luoqi and Dai, Jiao and Han, Jizhong and Li, Guanbin and Liu, Si},
  booktitle={Proceedings of the IEEE/CVF conference on computer vision and pattern recognition},
  pages={6966--6975},
  year={2024}
}

@article{oquab2024dinov2,
  title={DINOv2: Learning Robust Visual Features without Supervision},
  author={Oquab, Maxime and Darcet, Timoth{\'e}e and Moutakanni, Th{\'e}o and Vo, Huy and Szafraniec, Marc and Khalidov, Vasil and Fernandez, Pierre and Haziza, Daniel and Massa, Francisco and El-Nouby, Alaaeldin and others},
  journal={Transactions on Machine Learning Research Journal},
  pages={1--31},
  year={2024}
}

@inproceedings{in2024editsplat,
  title={Editsplat: Multi-view fusion and attention-guided optimization for view-consistent 3d scene editing with 3d gaussian splatting},
  author={Lee, Dong In and Park, Hyeongcheol and Seo, Jiyoung and Park, Eunbyung and Park, Hyunje and Baek, Ha Dam and Shin, Sangheon and Kim, Sangmin and Kim, Sangpil},
  booktitle={Proceedings of the Computer Vision and Pattern Recognition Conference},
  pages={11135--11145},
  year={2025}
}

\vfill

\end{document}